\documentclass{article}

\usepackage{microtype}
\usepackage{graphicx}
\usepackage{grffile}
\usepackage{booktabs} 

\usepackage{amsmath}
\usepackage{booktabs}
\usepackage[labelfont=bf,font=small]{caption}
\usepackage{algorithm}
\usepackage{algorithmic}
\usepackage{amsmath}
\usepackage{amssymb}
\usepackage{subcaption}
\usepackage{multirow}
\usepackage{blkarray}
\usepackage{xspace}
\usepackage{titling}
\usepackage{enumitem}
\usepackage{microtype}

\usepackage{xcolor}
\usepackage{listings}

\lstdefinelanguage{Python}{
 keywordstyle=\color{black}\bfseries,
 basicstyle=\small\fontencoding{T1}\fontfamily{fvm}\selectfont,
 identifierstyle=\color{black},
 sensitive=false,
 comment=[l]{\#},
 morecomment=[s]{/*}{*/},
 commentstyle=\color{green}\ttfamily,
 string=[s]{"}{"},
 showstringspaces=false,
 stringstyle=\color{violet}\ttfamily,
}
\usepackage{hyperref}

\newcommand{\tbw}{\texttt{2BW}\xspace}
\newcommand{\system}{PipeDream-\texttt{2BW}\xspace}
\newcommand{\pdtbw}{PipeDream-\texttt{2BW}\xspace}
\newcommand{\pdflush}{PipeDream-Flush\xspace}
\newcommand{\aggregation}{accumulation\xspace}

\usepackage[accepted]{icml2021}

\icmltitlerunning{Memory-Efficient Pipeline-Parallel DNN Training}

\begin{document}

\twocolumn[
\icmltitle{Memory-Efficient Pipeline-Parallel DNN Training}



\icmlsetsymbol{equal}{*}

\begin{icmlauthorlist}
\icmlauthor{Deepak Narayanan}{stan,equal}
\icmlauthor{Amar Phanishayee}{msr}
\icmlauthor{Kaiyu Shi}{ms}
\icmlauthor{Xie Chen}{ms}
\icmlauthor{Matei Zaharia}{stan}
\end{icmlauthorlist}

\icmlaffiliation{stan}{Stanford University}
\icmlaffiliation{msr}{Microsoft Research}
\icmlaffiliation{ms}{Microsoft}

\icmlcorrespondingauthor{Deepak Narayanan}{deepakn@cs.stanford.edu}

\icmlkeywords{Machine Learning, ICML}

\vskip 0.3in
]



\printAffiliationsAndNotice{\icmlEqualContribution} 

\begin{abstract}
Many state-of-the-art ML results have been obtained by scaling up the number of
parameters in existing models. However, parameters and activations for such
large models often do not fit in the memory of a single accelerator device;
this means that it is necessary to distribute training of large models over
multiple accelerators.  In this work, we propose \system{}, a system that
supports \emph{memory-efficient} pipeline parallelism.
\system{} uses a novel pipelining and weight gradient coalescing strategy,
combined with the double buffering of weights, to ensure high throughput, low
memory footprint, and weight update semantics similar to data parallelism. In
addition, \system{} automatically partitions the model over the available
hardware resources, while respecting hardware constraints such as memory
capacities of accelerators and interconnect topologies.
\system{} can accelerate the training of large GPT and BERT
language models by up to 20$\times$ with similar final model accuracy.
\end{abstract}

\section{Introduction}

In the quest to achieve higher accuracy across a range of tasks, DNN models
have grown in size, often by scaling up the number of parameters in existing
architectures~\cite{devlin2018bert, radford2018improving, radford2019language,
gpt3}.  It is challenging to train large models with billions of parameters.
Modern accelerators have limited memory, which means that the model parameters
and intermediate outputs that need to be in accelerator memory during training
might not fit on a single accelerator. One of the solutions researchers and
practitioners have turned to is model-parallel
training~\cite{dean2012deepbelief, chilimbi2014adam}, where a model is
partitioned over multiple accelerator devices. However, model parallelism, when
traditionally deployed, can either lead to resource
under-utilization~\cite{narayanan2019pipedream} or high communication overhead
with good scaling only within a multi-GPU server~\cite{shoeybi2019megatron},
and consequently an increase in training time and dollar cost.

\begin{figure*}[t!]
    \centering
    \begin{subfigure}[c]{\columnwidth}
        \centering
        \includegraphics[keepaspectratio=1.0,width=\columnwidth]{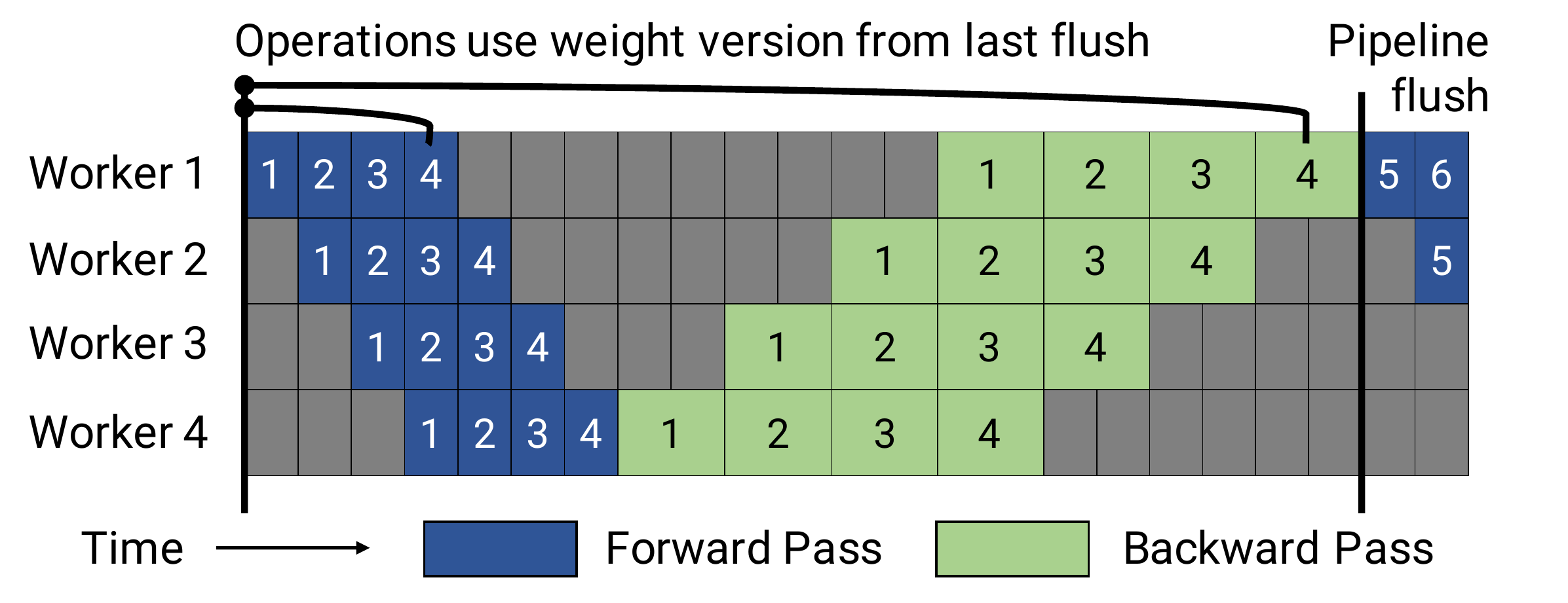}
        \caption{GPipe.}
        \label{fig:gpipe}
    \end{subfigure}
    \begin{subfigure}[c]{\columnwidth}
        \centering
        \includegraphics[keepaspectratio=1.0,width=\columnwidth]{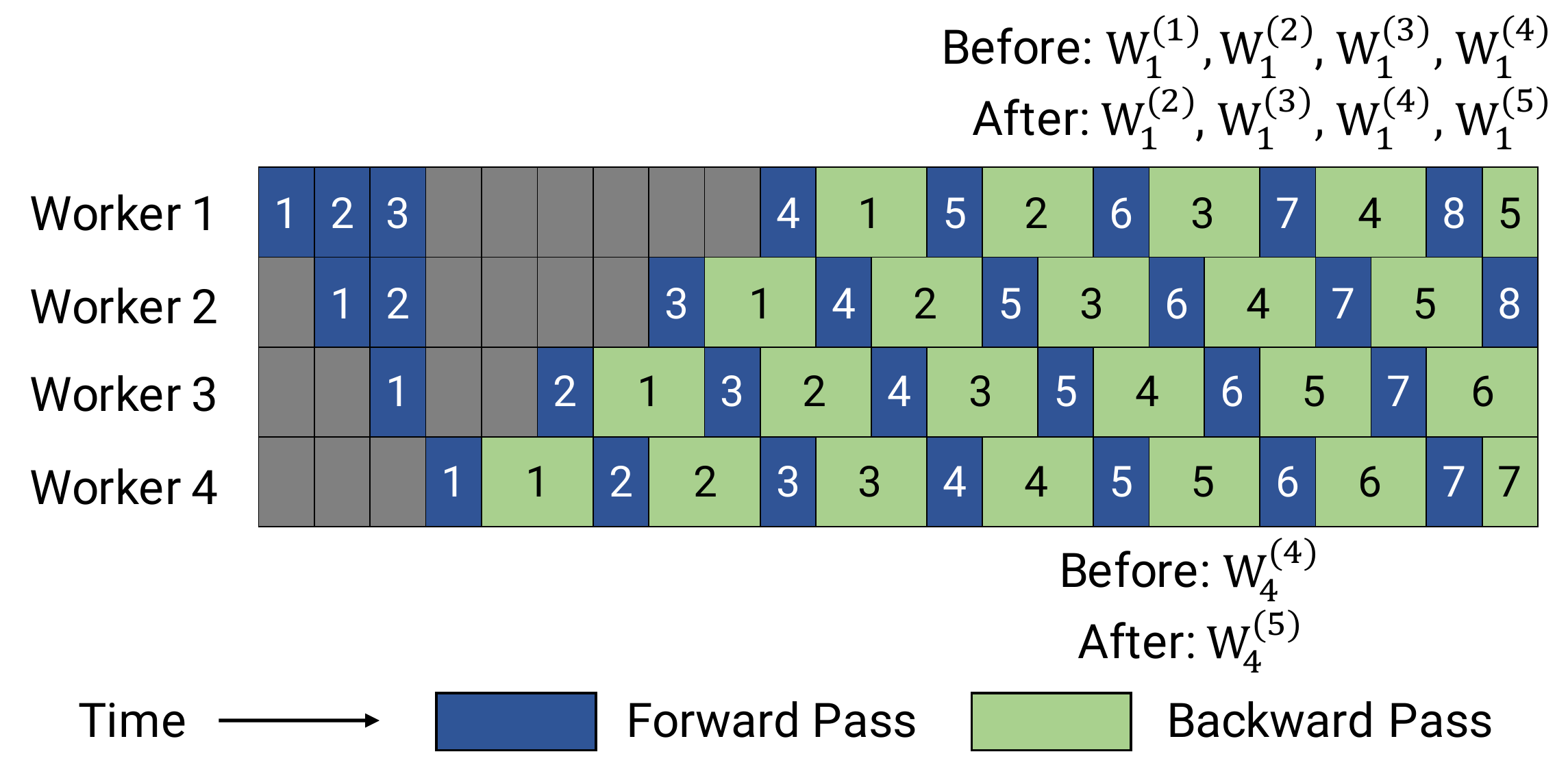}
        \caption{PipeDream.}
        \label{fig:pipedream}
    \end{subfigure}
    \vspace{-0.1in}
    \caption{
        Timelines of different pipeline-parallel executions. Without loss of generality,
        forward and backward passes are assumed to take twice as long as forward passes;
        forward passes are shown in blue and backward
        passes are shown in green. Numbers indicate microbatch ID, time is shown
        along $x$-axis, per-worker utilization is shown along the $y$-axis.
        GPipe maintains a single weight version, but periodically flushes
        the pipeline. PipeDream does not introduce periodic pipeline flushes,
        but maintains multiple weight versions.
    }
    \vspace{-0.1in}
\end{figure*}

Recent work has proposed \emph{pipelined} model parallelism to accelerate
model-parallel training. For example, GPipe~\cite{huang2019gpipe} and
PipeDream~\cite{harlap2018pipedream, narayanan2019pipedream} push multiple
inputs in sequence through a series of workers that each manage one model partition,
allowing different workers to process different inputs in parallel.
Na\"ive pipelining can harm model convergence due to inconsistent weight
versions between the forward and backward passes of a particular input.
Existing techniques trade off memory footprint and throughput in different ways
to avoid this. GPipe maintains a single weight version, but has periodic
\emph{pipeline flushes} where the pipeline is drained of inputs to update
weights (Figure~\ref{fig:gpipe}); these flushes limit overall throughput as
resources are idle. PipeDream does not periodically flush the pipeline but
stores multiple weight versions, which increases throughput but also increases
the memory footprint, making the training of large models infeasible due to
memory constraints.  Efficient training of large models requires an approach
with \emph{both} high throughput and low memory footprint.

Additionally, the performance of a pipeline-parallel system is dependent on how
DNN model operators are partitioned over workers. This is challenging for three reasons:
\begin{itemize}[itemsep=1pt,topsep=0pt,leftmargin=*]
\item \textbf{Memory Capacity Constraints:} Parameters and intermediate
activations associated with a model partition need to fit in the main device
memory of the accelerator.
\item \textbf{Heterogeneous Network Interconnects:} Training deployments today
feature heterogeneous network topologies, with higher-bandwidth links
between devices on the same server.
\item \textbf{Large Search Space for Operator Placement:} As model sizes
increase, splitting an operator graph becomes computationally expensive since
the number of distinct partitionings is exponential in the model size.
\end{itemize}
In this paper, we introduce \textbf{\system{}}, a system for efficient pipeline-parallel
training of DNN models with billions of parameters. \system{} achieves high
throughput \emph{and} low memory footprint using two key contributions.  First,
we propose \textbf{double-buffered weight updates} (\tbw), a technique
that reduces the memory footprint of training while avoiding pipeline flushes.
We leverage the fact that every input's generated gradient does not need to be
applied to weights immediately, and instead can be accumulated into a
``coalesced'' gradient to limit the number of weight versions maintained.
Instead of flushing the pipeline before using newly updated weights, \tbw uses
the new weights for inputs newly admitted into the pipeline,  while using the
previous weight version, called the \emph{shadow version}, for already
in-flight inputs.  This double buffering of weights at each worker yields a
pipelining scheme with higher throughput than GPipe (no pipeline flushes) and
better memory efficiency than PipeDream (2 weight versions, versus  worst case of
$d$ in PipeDream for a depth-$d$ pipeline). \tbw introduces a \emph{constant}
weight delay term of 1, consistent across stages, while updating weights
(weight update equation of $W^{(t+1)} = W^{(t)} - \nu \cdot \nabla
f(W^{(t-1)})$), which we show has empirically similar model convergence
to vanilla weight updates (\S\ref{sec:eval-convergence}).  We also present a variant of \tbw (called
\textbf{\pdflush}) that trades off throughput for even lower memory footprint and
vanilla semantics (weight update equation of $W^{(t+1)} = W^{(t)} - \nu \cdot
\nabla f(W^{(t)})$).

Second, \system{} provides a planning algorithm that yields effective
parallelization schemes for many of today's large model architectures.
\system{}'s \textbf{planner} partitions DNN operators over the available
workers while taking into account the memory capacities of the  accelerator
devices, and addresses the three challenges highlighted earlier.  \system{}'s
planner exploits the repetitive structure of large DNNs, e.g., transformer
layers in BERT~\cite{devlin2018bert}, to explore the space of schedules where
each stage in the pipeline is replicated \emph{equally}. This choice reduces
the size of the search space explored drastically compared to existing work
like PipeDream and FlexFlow~\cite{flexflow}, while still providing effective
model splits in practice. \system{}'s planner determines the size of each model
partition, batch size, and whether to use memory-saving optimizations like
\emph{activation recomputation}~\cite{chen2016training, griewank2000algorithm}.
\system{}'s planner considers the impact of these decisions on both throughput
and memory footprint, unlike PipeDream and FlexFlow. Finally, the planner tries
to ensure expensive communication stays on high-speed intra-server interconnects.

We find that the Adam optimizer with \tbw{} has a similar training loss
trajectory to vanilla Adam with the same batch size, with similar accuracy on
downstream finetuning tasks. \system{} achieves end-to-end speedups of
$1.3\times$ to $20\times$ for various GPT models compared to an
optimized model-parallel baseline. \system{} is up to 3.2$\times$ faster than
GPipe, and is able to train large transformer models that
vanilla PipeDream cannot fit in memory.

\section{Background} \label{sec:background}

In this section, we provide a brief overview of related techniques for
distributed training of DNN models.

\textbf{Data Parallelism.} Data parallelism is used to scale up model training.
With data parallelism~\cite{xing2015petuum}, every worker has a copy of the
entire model and the input dataset is sharded across workers. Data parallelism
cannot be used to train large models that do not fit on a single worker, but
can be used on smaller model partitions.

\textbf{Model Parallelism.} Model parallelism is used traditionally to train
large models that do not fit on a single worker. With model
parallelism~\cite{dean2012deepbelief, chilimbi2014adam}, the weight parameters
in a model are split over available workers, with intermediate activations and
gradients communicated across workers. Inter-layer model parallelism
underutilizes resources since at most a single worker is active at any point in
time. Tensor (intra-layer) model parallelism~\cite{shoeybi2019megatron} leads
to expensive all-to-all communication in the critical path, limiting the number
of model partitions to the number of GPUs in a single server.
FlexFlow~\cite{flexflow} shows how to split a model graph using model and data
parallelism, but still suffers from poor resource utilization when model
parallelism is used.

\textbf{Pipeline Parallelism.} To address the shortcomings of model
parallelism, recent work like PipeDream and
GPipe have proposed pipeline parallelism. With pipeline
parallelism, multiple inputs (instead of 1) are injected into a pipeline composed of inter-layer model partitions. This ensures that compute resources are better utilized.
However, naive pipelining can lead to weight version mismatches between forward
and backward passes for a particular input. Specifically, if weight updates are
immediately applied to the latest weight version, then an input might see
weight updates in the backward pass that it did not see in the forward pass,
leading to incorrect gradient computations.

GPipe maintains a single version of the model's weights. An input batch is
split into smaller \emph{microbatches}. Weight gradients are accumulated and
not applied immediately, and the pipeline is periodically \emph{flushed} to
ensure that multiple weight versions do not need to be maintained. GPipe
provides weight update semantics similar to data parallelism.
Figure~\ref{fig:gpipe} shows a timeline of GPipe execution. The periodic
pipeline flushes can be expensive, limiting throughput. One
way to mitigate this overhead is to perform additional accumulation within the
pipeline, but this is not always practical: a) at large scale factors, the minimum
supported batch size is large (proportional to the scale factor), and large
batch sizes affect convergence for all models (e.g.,
Megatron~\cite{shoeybi2019megatron} uses a batch size of 1024 for BERT and 512
for GPT with 512 GPUs), b) GPipe needs to maintain activation stashes
proportional to the batch size.

PipeDream uses a weight stashing scheme to ensure that the same weight version
is used in both the forward and backward passes for the same input
(Figure~\ref{fig:pipedream}). The total number of weight versions stashed is
$d$ in the worst case, where $d$ is the pipeline depth, which is too high for large
models. With PipeDream's default weight update semantics, weight updates at
each stage have different delay terms, and no \aggregation is performed within
the pipeline.

\section{\system{} System Design}

\system{} uses memory-efficient pipeline parallelism to train large models that
do not fit on a single accelerator. Its \emph{double-buffered weight update
(\tbw)} and \emph{flush} mechanisms ensure high throughput, low memory
footprint, and weight update semantics similar to data parallelism. \system{}
splits models into stages over multiple workers, and replicates each stage
an equal number of times (with data-parallel updates across replicas of the
same stage). Such \emph{parallel pipelines} work well for models where each
layer is repeated a fixed number of times (e.g., transformer models).

\subsection{Double-Buffered Weight Updates (\tbw)}
\label{sec:tbw}

\begin{figure*}[ht!]
    \centerline{\includegraphics[keepaspectratio=1.0,width=0.7\textwidth]{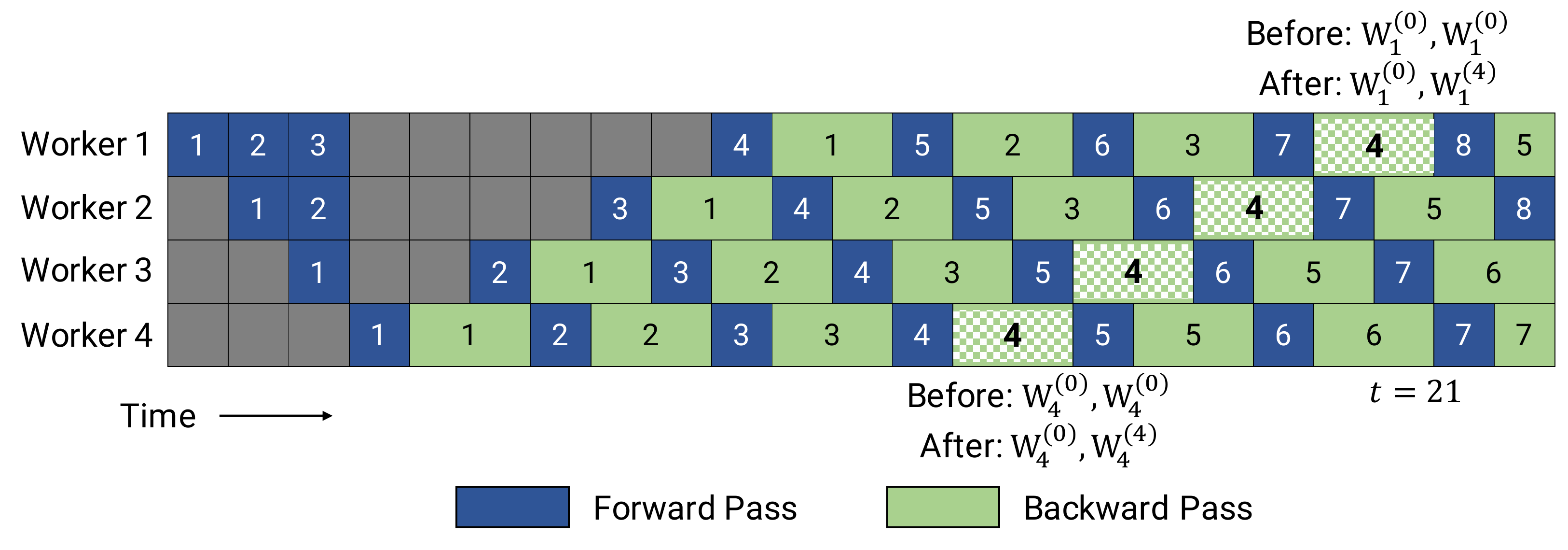}}
    \vspace{-0.1in}
    \caption{
       Timeline showing \system{}'s double-buffered weight update (\tbw) scheme
       with time along $x$-axis. \emph{Without loss of generality},
       backward passes are assumed to take twice as long as forward passes.
       \system{} only stashes two weight versions at every worker, reducing the
       total memory footprint while no longer requiring expensive pipeline stalls.
       $W^{(v)}_i$ indicates weights on worker $i$ with version $v$ (contains
       weight gradient generated from input $v$). New weight versions are
       generated in checkered green boxes; $W_4^{(4)}$ is first used for input 9's
       forward pass.
    }
    \label{fig:pipelining-timeline}
    \vspace{-0.2in}
\end{figure*}

\begin{figure}[t!]
    \centering
    \begin{subfigure}[c]{\columnwidth}
        \centering
        \includegraphics[keepaspectratio=1.0,width=0.95\columnwidth]{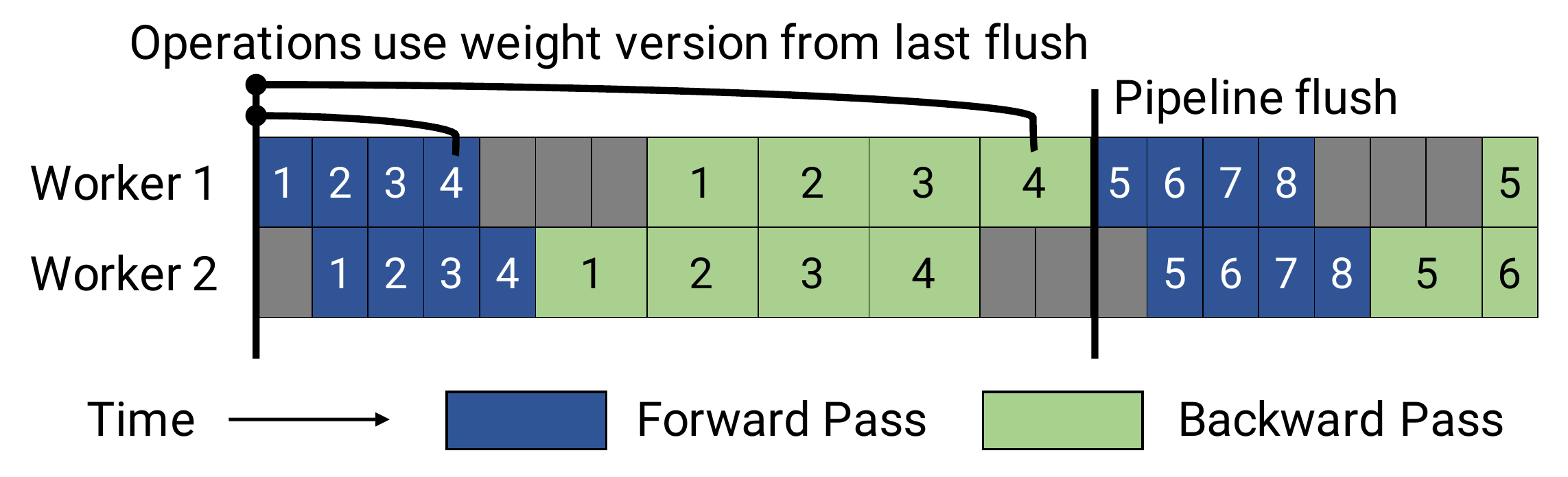}
        \caption{GPipe.}
    \end{subfigure}
    \begin{subfigure}[c]{\columnwidth}
        \centering
        \includegraphics[keepaspectratio=1.0,width=0.95\columnwidth]{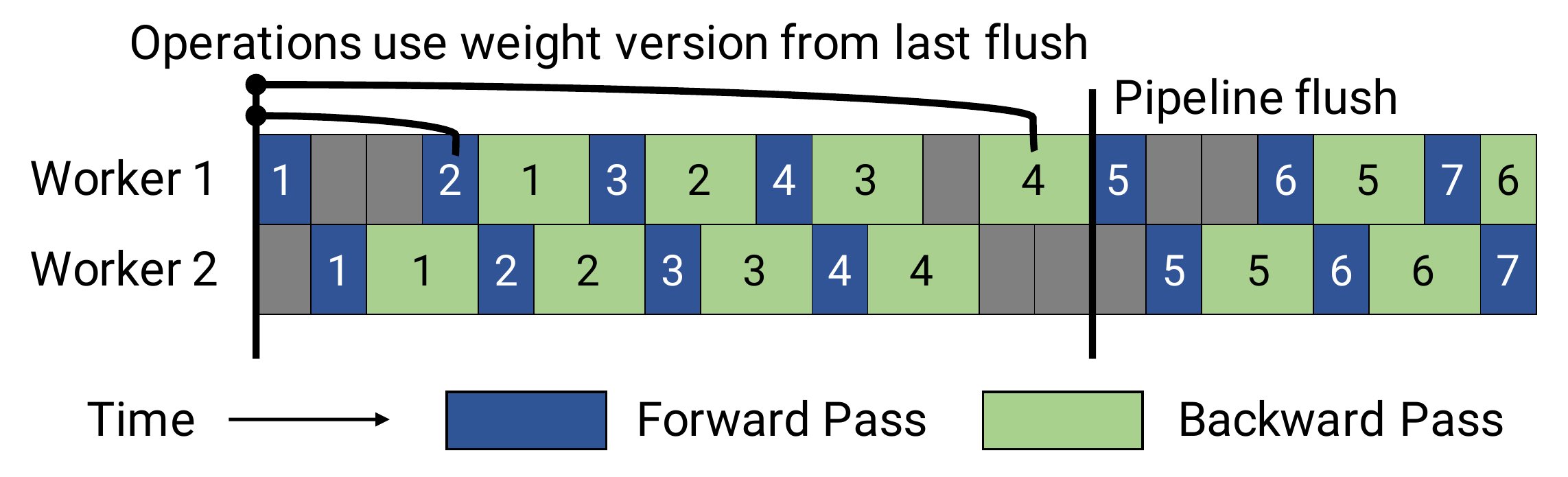}
        \caption{\pdflush.}
    \end{subfigure}
    \caption{
        Timelines of GPipe and \pdflush for 2 stages. Both GPipe and \pdflush
        use pipeline flushes; \pdflush alternates between forward and
        backward passes in steady state to keeping memory footprint low
        compared to GPipe by limiting activation stashes to only in-flight
        microbatches.
    }
    \label{fig:pipeline-flush-timeline}
    \vspace{-0.2in}
\end{figure}

\begin{figure}
    \centerline{\includegraphics[keepaspectratio=1,width=\columnwidth]{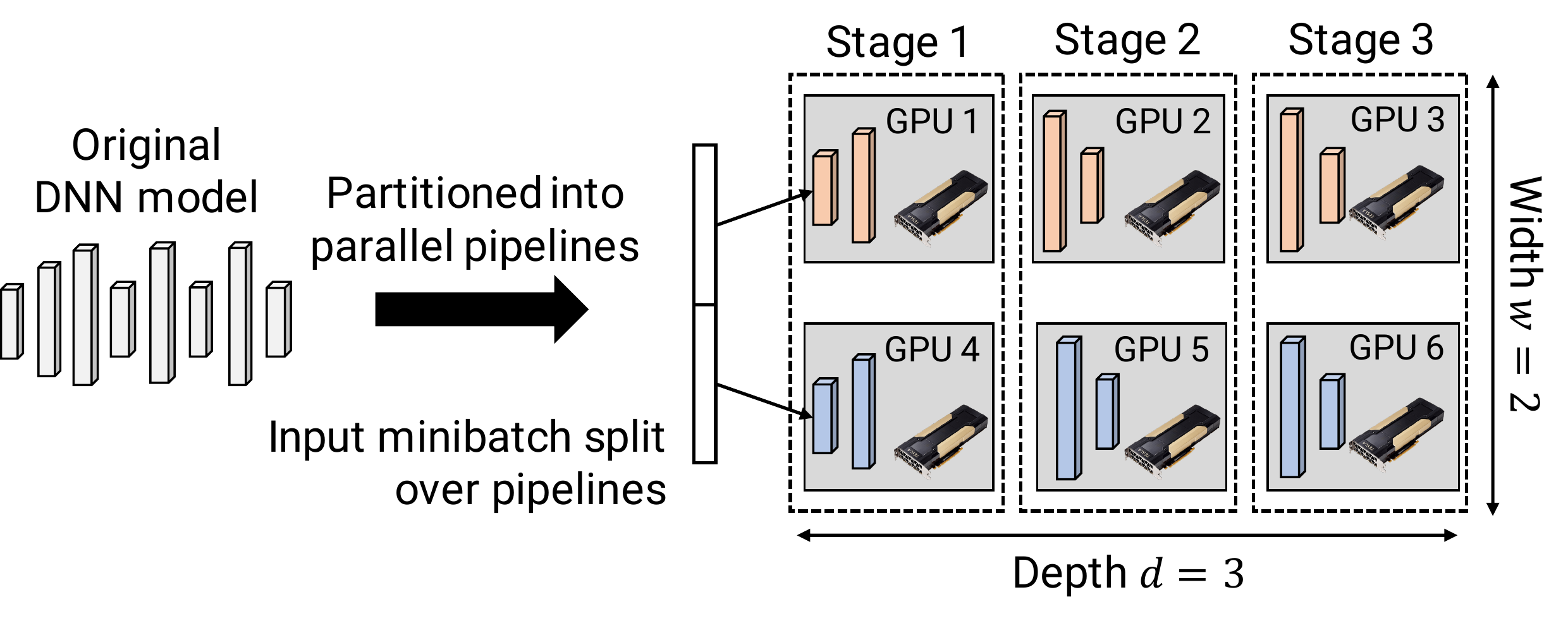}}
    \vspace{-0.1in}
    \caption{
       Example \system{} ($2, 3$) configuration. The model is partitioned into
       3 stages ($d=3$) and each pipeline is replicated twice ($w=2$). Each
       pipeline replica is shown in a different color.
    }
    \label{fig:fully-generic-parallelism-model}
    \vspace{-0.1in}
\end{figure}

\system{} uses a novel double-buffered weight update (\tbw) scheme in
conjunction with 1F1B scheduling~\cite{narayanan2019pipedream}, where each
worker alternates between forward and backward passes for different inputs, to
ensure that the same weight version is used in both the forward and the
backward pass for a particular input (Figure~\ref{fig:pipelining-timeline}).
\tbw has a lower memory footprint than PipeDream and GPipe, and also avoids
GPipe's expensive pipeline flushes.

Gradients are computed at the granularity of smaller \emph{microbatches}.  For
any input microbatch, \pdtbw{} uses the same weight version for an input's
forward and backward passes. Updates are accumulated over multiple microbatches
before being applied at the granularity of a batch, limiting the number of
weight versions generated and maintained. Figure~\ref{fig:pipelining-timeline}
shows an example timeline of \tbw. \system{} generates a new weight version
once every $m$ microbatches ($m \geq d$, the pipeline depth). For simplicity, we will initially assume that
$m=d$ ($d=4$ in Figure~\ref{fig:pipelining-timeline}). A new weight version
\emph{cannot} be used immediately. In particular, in-flight inputs cannot use
the newest weight version for their backward passes (for example, input 7 on
worker 3 at $t=21$), since the forward pass for these inputs was already
initiated using an older weight version on a different stage. Thus, newly
generated weight versions need to be buffered for future use.  However, the
total number of weight versions that need to be maintained is at most 2, since
the weight version used to generate a new weight version can immediately be
discarded (no future inputs that pass through that stage use the old weight
version any longer). For example, in Figure~\ref{fig:pipelining-timeline}, each
worker can discard $W_i^{(0)}$ once they are done processing the \emph{backward
pass} for input 8 since all subsequent inputs use a later weight version for
both their forward and backward passes.

The weight version a given input microbatch $k$ (1-indexed) uses is
$\max(\lfloor (k-1)/m \rfloor -1, 0)$, where $m$ is the number of microbatches in a
batch (4 in Figure~\ref{fig:pipelining-timeline}). This weight version is
the same for both the forward and backward passes for input $k$. $m$ can be any
number $\geq d$; additional gradient \aggregation (larger $m$) increases the
global batch size.

\textbf{Memory Footprint.} \pdtbw{} maintains 2 weight versions, and activation
stashes for all in-flight microbatches. The number of in-flight microbatches at any
stage is at most the pipeline depth ($d$).  With activation recomputation,
\pdtbw{}'s memory footprint can be decreased, since only input activations (as
opposed to the full intermediate activation) need to be maintained for all
in-flight microbatches.  With activation recomputation,
\pdtbw{}'s worst-case memory footprint is $\frac{2|W|}{d} + \frac{|A^\text{total}(b)|}{d} + d
|A^{\text{input}}(b)|$. $|W|$ is
the size of weight parameters for the full model, $|A^\text{total}(b)|$ is the size of
intermediate activations for microbatch size $b$ for the full model, and
$|A^{\text{input}}(b)|$ is the size of input activations for microbatch size
$b$ \emph{for a pipeline stage}.

In comparison, GPipe needs to checkpoint potentially a much larger number of
input activations -- proportional to the total number of microbatches
accumulated within the pipeline before applying a weight update ($m$).
With activation recomputation, GPipe's memory footprint with a
per-GPU microbatch size $b$ is $\frac{|W|}{d} + \frac{|A^\text{total}(b)|}{d} + m
|A^{\text{input}}(b)|$.  Since $|W| \ll
|A(b)|$ for even small $b$ for most models~\cite{jain2018gist}, the memory
savings from maintaining one fewer weight version is small.  To achieve high
throughput, GPipe must use a large value of $m$ to amortize away the cost of
pipeline flushes; at such high $m$, its memory footprint is higher than \pdtbw{}.
Additionally, due to its higher memory footprint,
GPipe must always use activation recomputation. Activation recomputation,
however, reduces throughput by about 33\%, and should be avoided if
possible.

\textbf{Semantics.} We can also formalize the semantics of \tbw.  For this
discussion, we assume an unreplicated pipeline with $d$ stages. If $b$ is the
per-GPU microbatch size, then gradients are averaged over $m$ microbatches; thus, the effective batch size is
$B=b\cdot m$.

We denote $W^{(t)}$ as the weight version after $t$ batches of size $B$.
$\nabla f(W)$ is the gradient averaged over the $B$ samples in the batch.
Vanilla minibatch SGD ($f$ is the loss function, $\nu$ is the learning rate)
then has the following weight update equation: $W^{(t+1)} = W^{(t)} - \nu \cdot
\nabla f(W^{(t)})$.  \tbw's weight update semantics (with a delay term of 1
\emph{across} all stages) are almost unchanged: $$W^{(t+1)} = W^{(t)} - \nu
\cdot \nabla f(W^{(t-1)}).$$ We show that this delay term does not affect model
convergence significantly in \S\ref{sec:eval-convergence}. Intuitively, the
parameters of the model do not change significantly across single iterations,
so $W^{(t)} \approx W^{(t-1)}$.  The semantics with a replication factor
greater than 1 is similar, with the batch size multiplied by the number of replicas
(as with regular data parallelism). Other momentum-based optimizers such as Adam can
be similarly analyzed (momentum term uses a weight gradient computed on a
1-stale weight version instead of latest version).  Extra shadow variables are
not needed. For example, $m_t$ in minibatch SGD with momentum can be computed
as (ignoring bias corrections) $m_t=\beta \cdot m_{t-1}+(1-\beta) \cdot \nabla
f(W^{(t-1)})$.  The final weight update equation is then $W^{(t+1)}=W^{(t)}-\nu
\cdot m_t$.

\subsection{Weight Updates with Flushes (\pdflush)}

We also propose a second memory-efficient pipeline schedule called \pdflush.
It has lower memory footprint than \tbw and vanilla optimizer semantics, at the
cost of lower throughput. This schedule reuses the 1F1B schedule from
PipeDream~\cite{narayanan2019pipedream}, but maintains a single weight version
and introduces periodic pipeline flushes to ensure consistent weight versions
across weight updates. Timelines for \pdflush and GPipe with 2 pipeline stages
are shown in Figure~\ref{fig:pipeline-flush-timeline}.

\textbf{Memory Footprint.} With \pdflush, the total number of in-flight
``active'' input activations is less than or equal to the pipeline depth,
giving it lower memory footprint than GPipe, which has to maintain input
activations proportional to the number of microbatches over which gradients are averaged ($m$).
\pdflush's memory footprint is also lower than \pdtbw{} since it only needs to
maintain a single weight version (versus 2 with \pdtbw{}).

\textbf{Semantics.} Periodic pipeline flushes ensure that weight updates can be
performed with gradients computed using the latest weight version.  This
results in weight updates of the form $W^{(t+1)} = W^{(t)} - \nu \cdot \nabla
f(W^{(t)})$. We compare \tbw's statistical efficiency (rate of model convergence) to the vanilla
semantics of \pdflush, GPipe, and data parallelism, in
\S\ref{sec:eval-convergence}.

\subsection{Equi-replicated Stages (Parallel Pipelines)}

\system{} executes DNN training using a hybrid parallelization scheme which
combines data and model parallelism with input pipelining. Since large deep
models today feature extremely repetitive structures, with the same block
repeated multiple times, a simple way of load balancing computation and
communication involves breaking up a model into stages with an equal number of
blocks and replication factors. Model training in \system{} can thus be thought
of as a collection of parallel pipelines
(Figure~\ref{fig:fully-generic-parallelism-model}), where inputs and
intermediate output activations within a pipeline do not ever need to be sent to workers
responsible for a different pipeline. Intermediate activations and gradients
can be communicated \emph{within} a pipeline using point-to-point communication
primitives, such as \lstinline{send} and \lstinline{recv}.  As with PipeDream,
weight gradients need to be aggregated across stage replicas in different
pipelines.  Figure~\ref{fig:fully-generic-parallelism-model} shows an example: each model copy is split across 3 workers (number
of stages or depth, $d=3$), and each stage is replicated twice (number of pipelines
or width, $w=2$).  Stage replicas can be placed on the same server so that
expensive all-reduce updates are between GPUs on the same server with
high-bandwidth interconnects.

\section{Planner}
\label{sec:planner}

\system{}'s \emph{planner} determines how to split a model over the available
compute devices by exhaustively searching over the \emph{reduced} search space
of all possible parallel-pipeline configurations. The planner also determines
whether memory-saving optimizations should be deployed, and the per-GPU
microbatch size and degree of gradient \aggregation, given a maximum
\emph{safe} global batch size verified to not compromise model convergence
(e.g., determined from past hyperparameter sweeps without pipelining).

\system{}'s planner uses a cost model for the compute times and memory
footprints of individual blocks in the model. Time and memory cost functions
allow \system{} to reason about the impact of pipeline width / depth and memory-saving
optimizations (such as activation recomputation) on throughput and memory
footprint. For example, a deeper configuration has additional memory capacity,
allowing for a larger maximum per-GPU microbatch size; this can increase the
arithmetic intensity (number of floating point operations performed per memory
load) of kernels~\cite{jouppi2017datacenter}, and consequently throughput.
Communication times for tensors can be estimated by dividing the size of the
tensor by the respective bandwidth. Expensive communication (e.g., large
tensors, or all-reduce communication needed to coalesce weight gradients across stage replicas) can be placed on high-bandwidth
links within the server by orienting pipelines appropriately.

Profiling for cost modeling can be done in two ways: end-to-end for each distinct configuration,
or extrapolating from an individual block's measurements.  End-to-end profiling
is cheap (2 to 3 minutes per configuration), which means total profiling time
is still a couple of hours (compared to the days to weeks needed for model
training). Optimal configurations can be reused for a given server and model
deployment. We describe how
per-block time and memory measurements can be extrapolated in Appendix \S\ref{sec:planner_appendix} -- this is even
cheaper, but provides less accurate cost estimates. The
highest-throughput-configuration is chosen that also fits within the memory
capacity of the target accelerators.


\subsection{Activation Recomputation} Activation recomputation is a common
technique~\cite{huang2019gpipe, chen2016training, griewank2000algorithm} that
trades off extra computation for a lower memory footprint. With activation
recomputation, activation stashes are not left materialized on the device
between forward and backward passes; instead, only \emph{input} activations on
each stage are stashed, and the remaining activations needed in the backward
pass are recomputed when required by re-running the forward pass. Activation
recomputation trades off extra computation for a lower memory footprint.

Activation recomputation is useful for two reasons: it can enable larger
per-GPU microbatch sizes to fit in memory, which can improve device throughput
by increasing the arithmetic intensity of kernel. 
It can also enable the training of large models. Concretely, in some cases, the
target accelerator device does not have sufficient memory capacity to store
full activation stashes for all in-flight microbatches. This is especially true
for deep pipelines, since the number of in-flight inputs is proportional to the
depth of the pipeline~\cite{narayanan2019pipedream}.

\subsection{Partitioning Algorithm} Putting it all together, given a total
memory capacity $M$, \system{}'s planner first determines the largest per-GPU
microbatch size that fits on a given worker (and the corresponding throughput)
with and without each memory-savings optimization deployed using a memory cost
function. The partitioning algorithm also verifies that the resulting global
batch size is lower than the maximum safe batch size $B$ . Each memory-savings
optimization can be integrated into \system{}'s planner by specifying a
corresponding throughput and memory cost function.

\system{}'s planner then sweeps all $(w, d)$ values to determine the best
pipeline configuration for a given model and hardware deployment.
Configurations with memory footprint higher than the memory capacity $M$ of the
device (modeled by the
\textsc{memory}(.) cost function) are discarded.  Gradient \aggregation can be used to increase the batch
size to $B$.  The partitioning algorithm aims to pick a configuration that has
a high compute-to-communication ratio, while accounting for the communication
time across stages in the same pipeline and across replicated stages (modeled by the
\textsc{throughput}(.) cost function). The full algorithm is shown
in Appendix \S\ref{sec:planner_appendix}.

\section{Evaluation}
\label{sec:evaluation}

In this section, we show that the Adam optimizer with \tbw has similar
semantics to vanilla Adam, and that \pdtbw{} and \pdflush{} are able to train
large models faster than existing model-parallel approaches including
Megatron~\cite{shoeybi2019megatron}, and existing pipelining approaches like
GPipe~\cite{huang2019gpipe}.

\textbf{Hardware.} We show results on two different hardware setups on AWS:
eight 8$\times$V100 servers (64 GPUs) with \lstinline{NVLink} and 16GB of
per-GPU memory, and a single 8$\times$V100 server. We use
\lstinline{p3.16xlarge} instances.

\textbf{Implementation.} Our implementation uses PyTorch and is adapted from
the Megatron repository~\cite{megatronrepository}; we verified that
single-worker performance  with this implementation achieves about 45 TFLOPS on
a 355M-parameter GPT model and is competitive with existing state-of-the-art
open source implementations from NVIDIA~\cite{nvidiabenchmark}. All results
shown are with mixed precision.

\textbf{Models.} We evaluate \system on BERT~\cite{devlin2018bert} and
GPT~\cite{radford2019language}, large transformer-based language models used
for a number of NLP applications.  In particular, most of our experiments are
performed with GPT models with 1.3, 2.2, and 3.9 billion parameters, with
similar layer dimensions to those used in the Megatron
paper~\cite{shoeybi2019megatron}.

\textbf{Baselines.} We compare \system{} to two types of baselines: (a) model
parallelism without pipelining (tensor model parallelism used in Megatron, and
inter-layer model parallelism); and (b) GPipe (we extend GPipe to
use parallel pipelines, and refer to this \emph{enhanced} version as GPipe in
the rest of this paper), which performs pipeline
parallelism. We do not compare to PipeDream or data parallelism for the entire
model since they cannot fit the above models in memory when using 16-GB V100
GPUs. With 64 GPUs, we use data parallelism \emph{across stages} to scale up
training.

\textbf{Main Takeaways.} We make the following observations:
\begin{itemize}[itemsep=1pt,topsep=0pt,leftmargin=*]
\item \textbf{Quality of Convergence:} \tbw weight update semantics
yield pre-trained models which produce \textbf{comparable accuracy}
on downstream finetuning tasks to vanilla Adam (GPipe and \pdflush) with the
same batch size.
\item \textbf{Comparison to Model Parallelism:} \system{} is able to
train a 3.8 billion-parameter GPT model up to \textbf{20$\times$} faster
compared to non-pipelining approaches.
\item \textbf{Comparison to Other Pipelined Approaches:} \system{} is up to
\textbf{3.2$\times$} faster than GPipe.
\end{itemize}

\subsection{Quality of Convergence of \tbw}
\label{sec:eval-convergence}

\begin{figure}[t!]
    \centering
    \begin{subfigure}[c]{\columnwidth}
        \includegraphics[keepaspectratio=1.0,width=0.49\columnwidth]{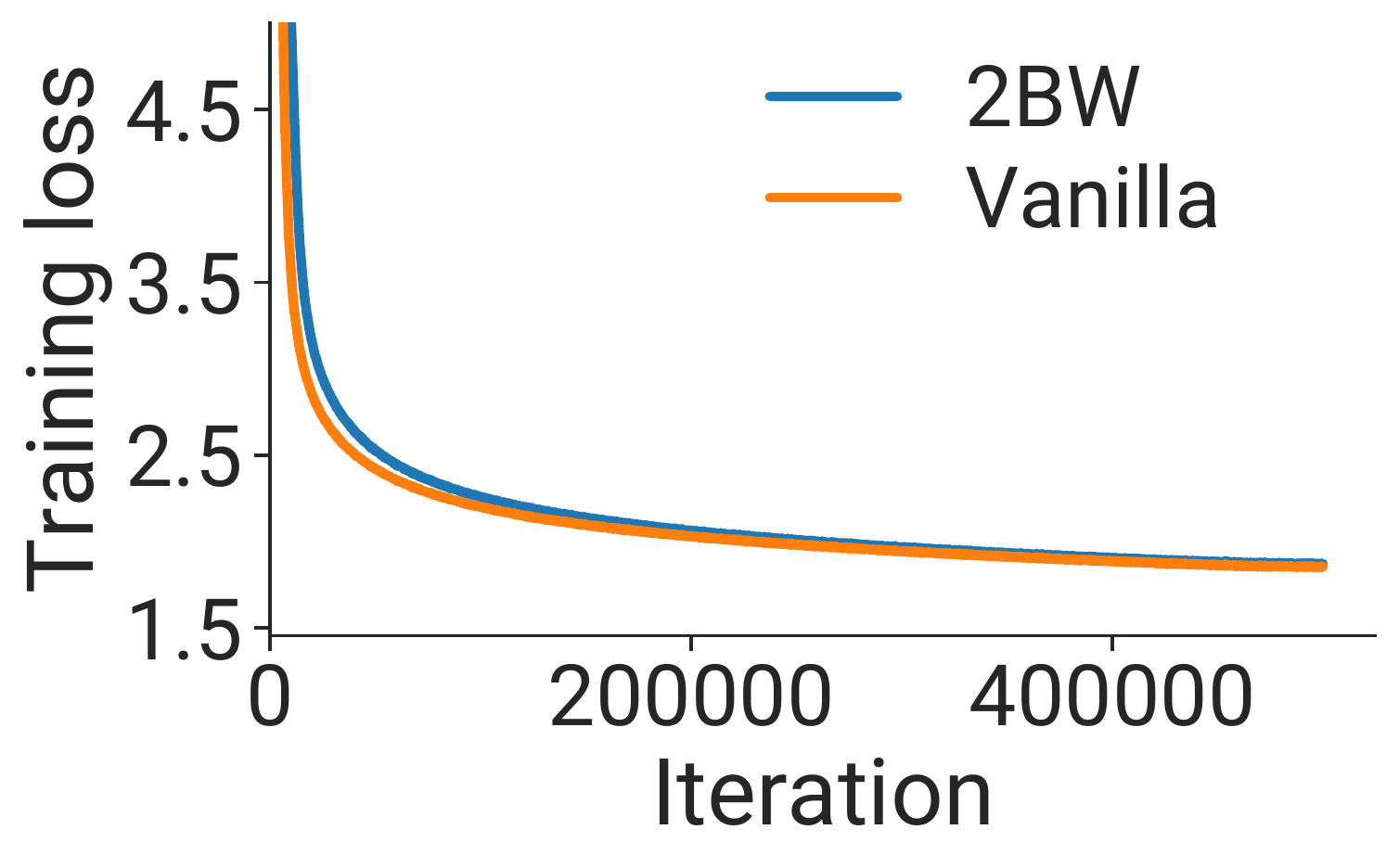}
        \includegraphics[keepaspectratio=1.0,width=0.49\columnwidth]{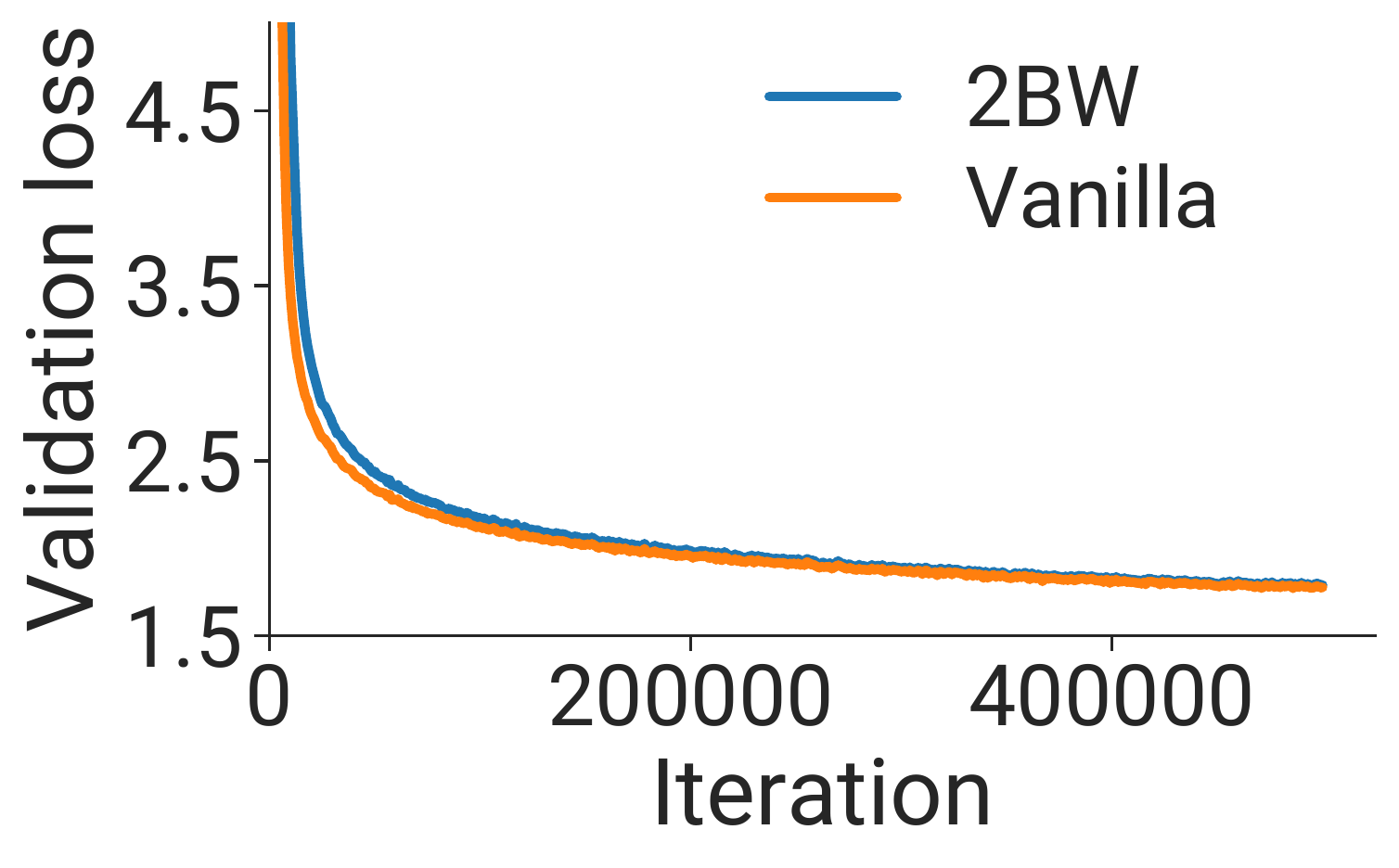}
        \caption{BERT, 355M (batch size = 1024).}
    \end{subfigure}
    \begin{subfigure}[c]{\columnwidth}
        \includegraphics[keepaspectratio=1.0,width=0.49\columnwidth]{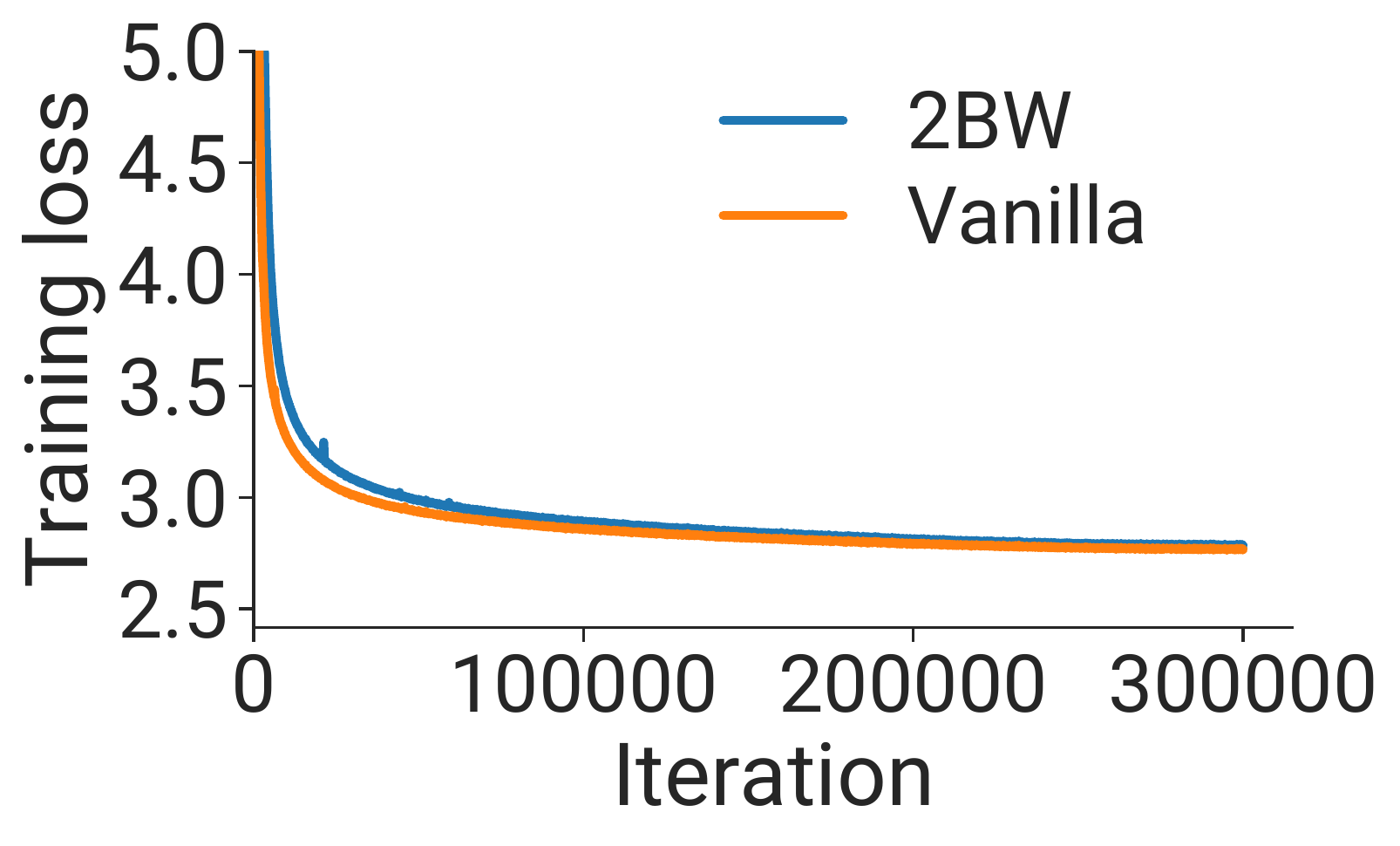}
        \includegraphics[keepaspectratio=1.0,width=0.49\columnwidth]{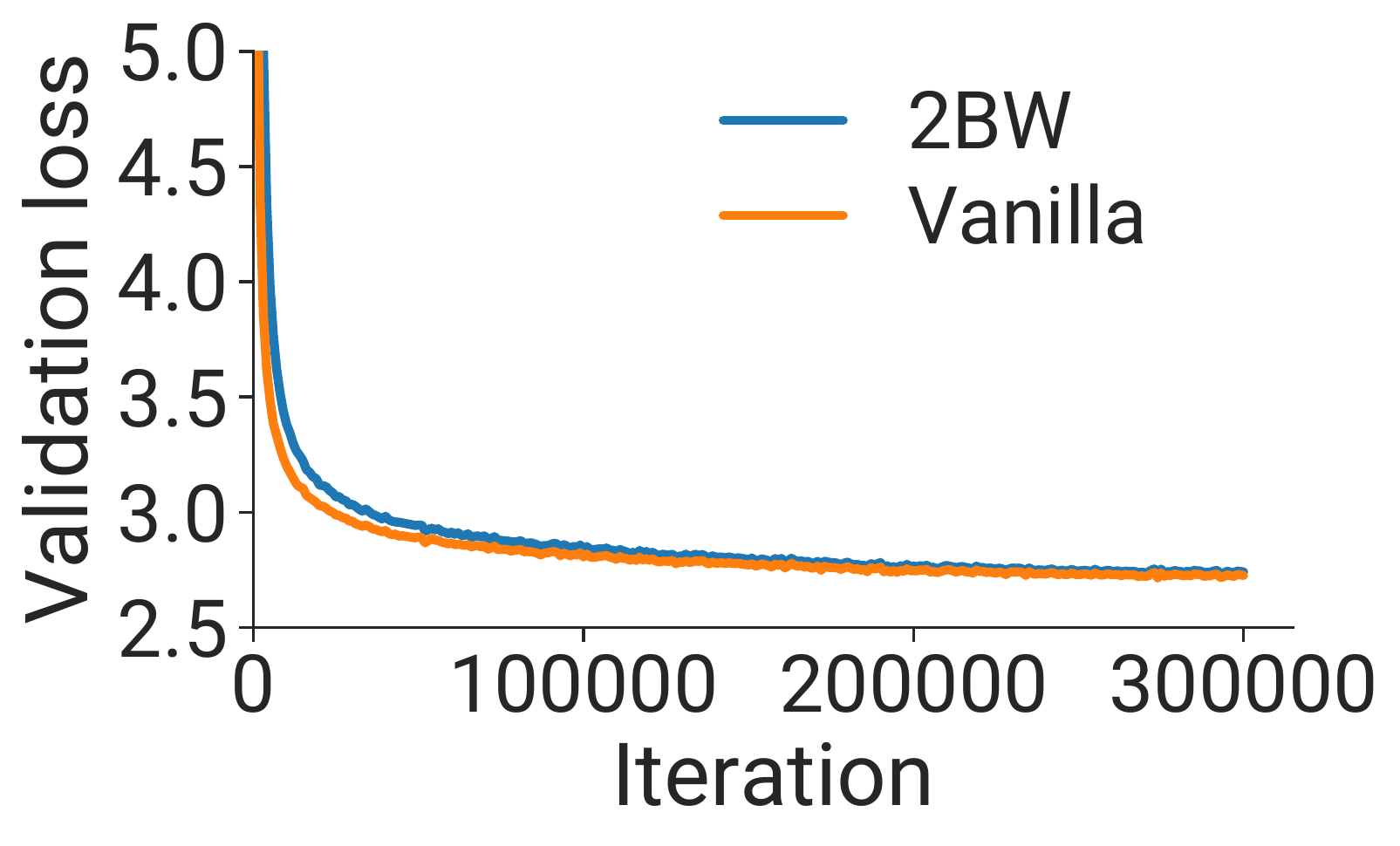}
        \caption{GPT, 355M (batch size = 512).}
    \end{subfigure}
    \vspace{-0.1in}
    \caption{
        Training and validation loss when pre-training BERT and GPT models with
        vanilla Adam and Adam with \tbw.
    }
    \vspace{-0.1in}
    \label{fig:convergence}
\end{figure}

\begin{table}[t]
\footnotesize
\centering
\begin{tabular}{ccccc}
\toprule
Task  &  Metric        &  Vanilla &  Vanilla (90\%)      & \tbw \\
\toprule
MNLI  &  Overall Acc.        & 87.77\% & N/A & 87.82\% \\
RACE  &  Overall Acc.   & 80.06\% & 79.30\% & 79.48\% \\
\bottomrule
\end{tabular}
\caption{Comparison of BERT models pre-trained with vanilla (all and 90\% of iterations)
         and \tbw optimizers on finetuning tasks.}
\label{table:finetuning}
\vspace{-0.2in}
\end{table}


We pre-trained 355M-parameter BERT and GPT models with vanilla Adam
and Adam with \tbw; we then finetuned the resulting BERT models. We note that
GPipe, \pdflush, and DP have identical semantics, and hence are equivalent
baselines (``Vanilla''). To provide a fair comparison, we use the \emph{same} hyperparameters, including batch size, used by Megatron~\cite{shoeybi2019megatron} to train these BERT and GPT models.
For BERT, we use a batch size of 1024, and for GPT, we
use a batch size of 512.
We use the Adam optimizer with standard hyperparameters
(learning rate of $10^{-4}$ with initial warmup and subsequent linear decay,
maximum sequence length of 512), and mixed precision.
We used the OpenWebText dataset~\cite{openwebtext} for pretraining. 
Figure~\ref{fig:convergence} shows the training and validation loss for the two
models.  The training and validation losses for the \tbw runs track the vanilla
runs almost identically after the first 100k iterations (when the model is
changing more rapidly and the delay term matters more).

To further validate the quality of the pre-trained model, we finetuned
the pre-trained vanilla and \tbw BERT models on downstream MNLI and RACE
tasks~\cite{wang2019glue,lai2017race}. Both pre-training and fine-tuning were
performed with the same hyperparameter and training setups, and we did not
perform hyperparameter tuning for either -- our goal here is to show that \tbw
has nearly identical semantics to the corresponding vanilla optimizer. As shown
in Table~\ref{table:finetuning}, the accuracy on each of these tasks is similar
after finetuning. We also evaluated the vanilla and \tbw GPT models on the
Wikitext-103 test dataset and got similar test perplexities (19.28 vs.  19.56);
test perplexities match \emph{exactly} when ``Vanilla'' is run for 20\% fewer
iterations.

\begin{figure}[t!]
    \centering
        \begin{subfigure}[c]{\columnwidth}
            \centering
            \includegraphics[keepaspectratio=1.0,width=0.82\columnwidth]{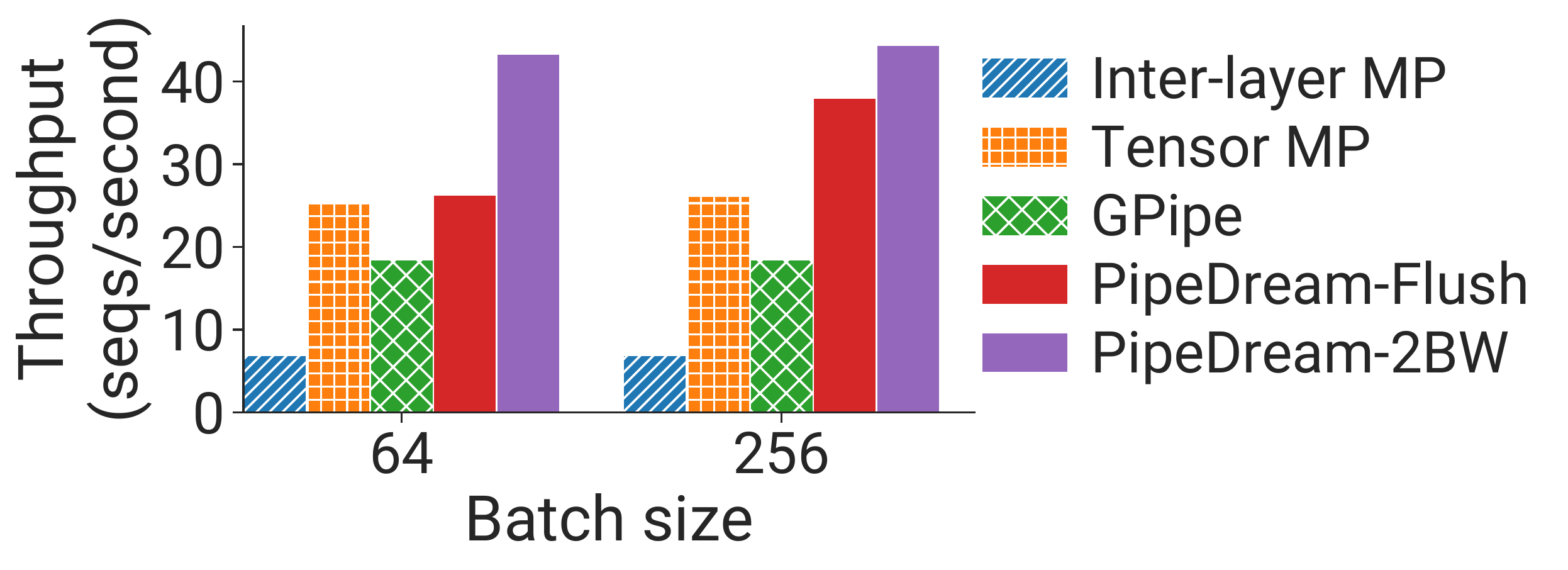}
            \vspace{-0.1in}
            \caption{GPT, 2.2B, 8-way model parallelism (8$\times$V100s).}
        \end{subfigure}
        \begin{subfigure}[c]{\columnwidth}
            \centering
            \includegraphics[keepaspectratio=1.0,width=0.82\columnwidth]{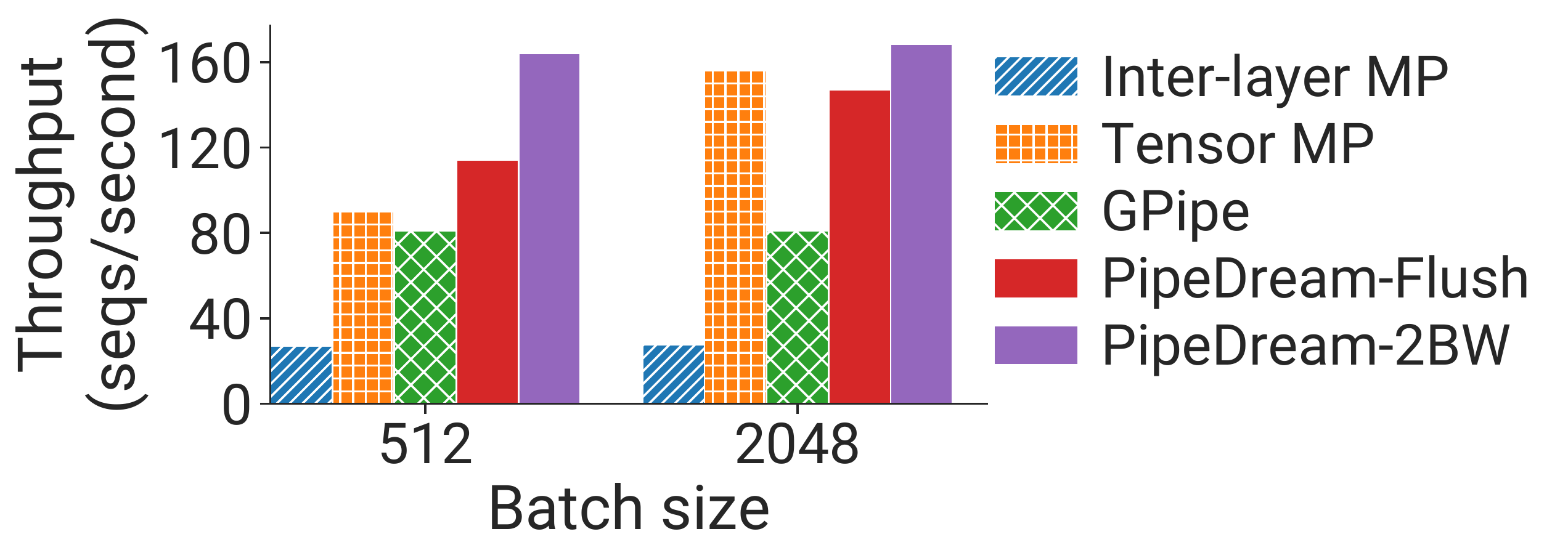}
            \vspace{-0.1in}
            \caption{GPT, 2.2B, 8-way model parallelism (64$\times$V100s).}
        \end{subfigure}
        \begin{subfigure}[c]{\columnwidth}
            \centering
            \includegraphics[keepaspectratio=1.0,width=0.82\columnwidth]{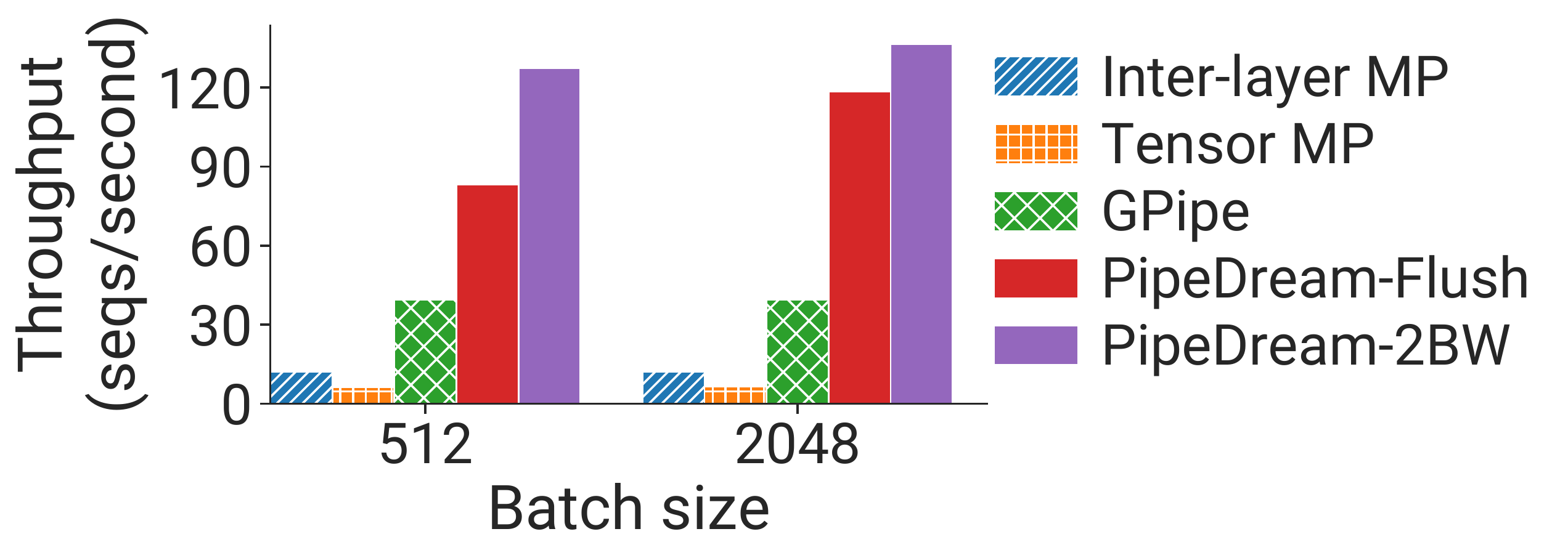}
            \vspace{-0.1in}
            \caption{GPT, 3.8B, 16-way model parallelism (64$\times$V100s).}
        \end{subfigure}
    \vspace{-0.1in}
    \caption{
       Throughput of various systems for different batch sizes for
       GPT models, using 8$\times$16GB-V100 servers.
    }
    \label{fig:throughput_vs_other_pipelining_approaches}
    \vspace{-0.2in}
\end{figure}

\subsection{Throughput}
\label{sec:evaluation_throughput_comparison}

Figure~\ref{fig:throughput_vs_other_pipelining_approaches} shows the
throughputs of various \system{}, \pdflush, and baseline configurations using 8 and 64
V100s with a sequence length of 512 for various large GPT models. Results with
BERT models are similar and included in Appendix \S\ref{sec:evaluation_bert}. We compare to two
different forms of model parallelism, as well as GPipe. Data parallelism is
not a viable baseline for these large models due to its high memory
overhead. In these experiments, we use activation recomputation, and the
largest per-GPU microbatch size that fits on the 16-GB V100 GPUs. We use the
best configuration recommended by \system{}'s planner for all comparisons:
8-deep configurations for the model with 2.2 billion parameters, and 16-deep
configurations for the model with 3.8 billion parameters.  For each model, we
show two different batch sizes to show the impact of batch size on throughput
for approaches that use periodic flushes. 

\textbf{Model Parallelism without Pipelining:} We compare against two model
parallelism approaches: tensor model parallelism used by
Megatron~\cite{shoeybi2019megatron} where each layer is divided among all
model-parallel workers, and inter-layer model parallelism where layers are
sharded over the workers but inputs are not pipelined. On a single node, \system{} is faster than tensor MP
by \textbf{1.3$\times$}. This grows to \textbf{20$\times$} on 64 GPUs for the
model with 3.8 billion parameters, when the all-to-all communication used by
tensor MP needs to be performed across servers, which is expensive using AWS instances (bandwidth across multi-GPU servers is much lower than
the bandwidth within server). Compared to inter-layer MP,
pipelining with flushes increases throughput by up to \textbf{4.1$\times$} for
small batch sizes, and by up to \textbf{5.3$\times$} for large batch sizes, on the
2.2-billion model. \tbw is up to \textbf{6.1$\times$} faster than
inter-layer MP.

\textbf{GPipe:} \system{} outperforms corresponding GPipe configurations at the
same global batch size by up to \textbf{3.2$\times$} due to the lack of
periodic pipeline flushes. GPipe natively has high memory footprint due to a
large number of activation stashes: consequently, the maximum number of
microbatches it can admit is small, leading to a larger pipeline bubble and
\textbf{2.1$\times$} worse throughput than \pdflush at low batch sizes, and
\textbf{3$\times$} at high batch sizes.

\textbf{\pdflush and \pdtbw:}
Figure~\ref{fig:throughput_vs_other_pipelining_approaches} also compares
\system and \pdflush for two different batch sizes with different numbers of microbatches
over which gradients are averaged ($m = d \cdot g$) within the pipeline. At low batch size, \system is
up to \textbf{1.6$\times$} faster. With more gradient accumulation (batch size
of 2048), this speedup drops to \textbf{15\%}. However, high $g$ is not always
practical.  Both \pdflush and \system{} have weight updates with a batch size
of $b \cdot w \cdot d \cdot g$, where the total number of workers is $w \cdot
d$. For a large number of workers ($\gg64$), the batch size is high even with
$g=1, m=d$, making additional gradient accumulation infeasible (batch size cannot
scale to $\infty$ without affecting model convergence).  Indeed, systems like
Megatron~\cite{shoeybi2019megatron}, that train large transformer models using
512 GPUs, show state-of-the-art results across tasks using a global batch size
$\leq 1024$.

\begin{figure}[t!]
    \centering
    \includegraphics[keepaspectratio=1.0,width=0.84\columnwidth]{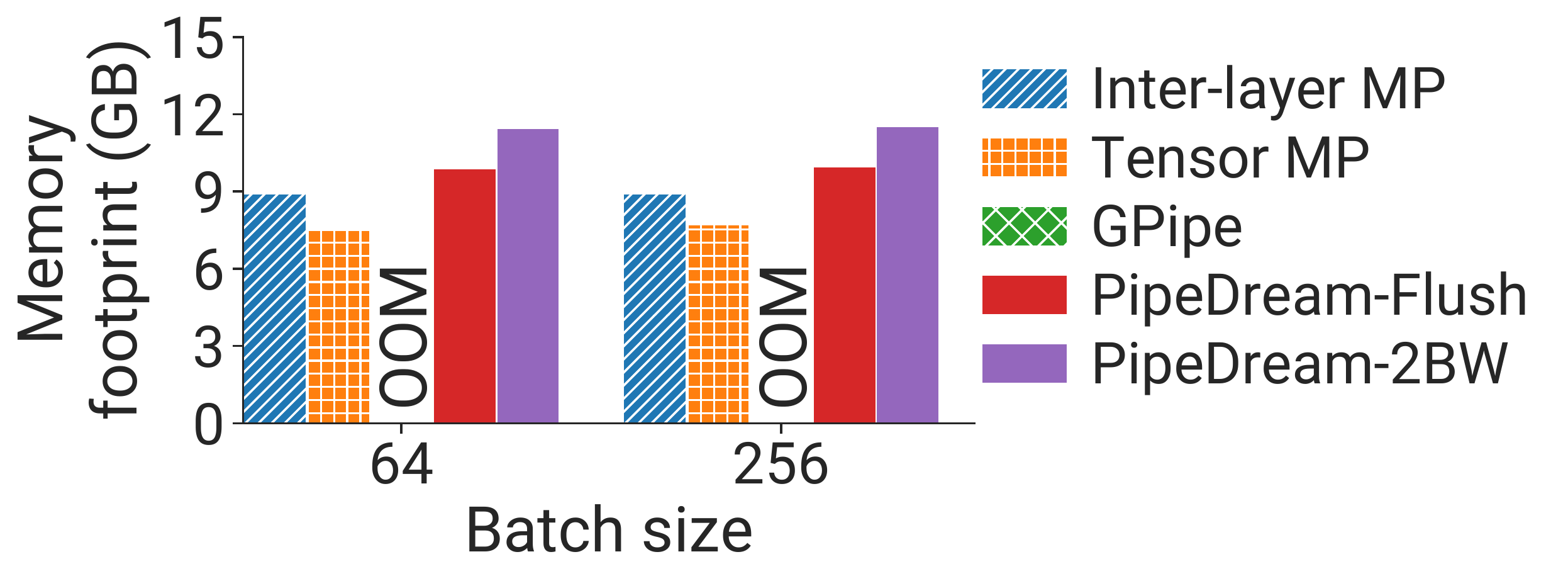}
    \vspace{-0.2in}
    \caption{
        Worst-case memory footprint (in GB) of various systems
        with 8 V100 GPUs for a GPT model with 2.2 billion parameters.
    }
    \vspace{-0.1in}
    \label{fig:memory_footprint}
\end{figure}

\begin{figure}[t!]
    \centering
    \includegraphics[keepaspectratio=1.0,width=0.78\columnwidth]{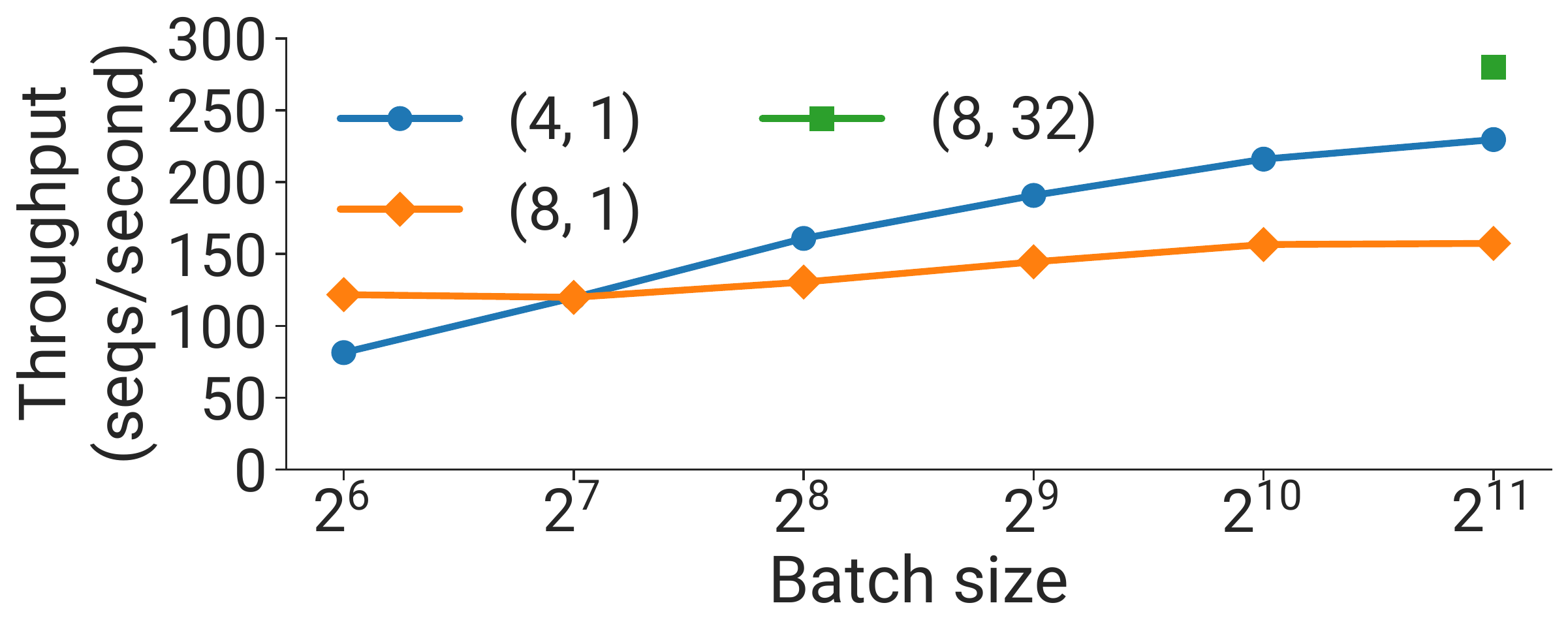}
    \vspace{-0.15in}
    \caption{
       Throughput of two \system{} configurations vs. global batch size
       for a 1.3-billion parameter GPT model using 64 V100 GPUs. The legend
       shows $(d, b)$: the number of pipeline-parallel stages and the microbatch
       size.
    }
    \vspace{-0.15in}
    \label{fig:throughput_vs_batch_size}
\end{figure}

\subsection{Memory Footprint}
We measured the worst-case memory footprint of different systems on a GPT model,
shown in Figure~\ref{fig:memory_footprint}.  GPipe runs out of
memory at a batch size of 64, due to a larger number of activation stashes
from its all-forward-all-backward schedule, even with activation
recomputation (worst case of $m$ input activation stashes with activation recomputation, compared to $d$ for \pdflush).  \pdflush has a slightly higher memory footprint compared to
inter-layer model parallelism,
since it needs to maintain activation stashes for more in-flight microbatches.
\system{} has a higher memory footprint than \pdflush due to an additional
weight version (but still lower than GPipe's).

\subsection{Planning Decisions}
\label{sec:evaluation_bagpipe_configuration_comparison}

In this sub-section, we analyze the implications of pipeline depth and width on
performance. We show experiments demonstrating the impact of activation
recomputation on performance in Appendix \S\ref{sec:evaluation_bagpipe_configuration_comparison_activation_recomputation}.
Figure~\ref{fig:throughput_vs_batch_size} shows the throughputs of two
\system{} configurations for different batch sizes. We highlight relevant
takeaways below.

\textbf{Inter-Stage Communication:} As the global batch size increases with
gradient \aggregation, throughput for each configuration increases due
to less communication across stage replicas. This is especially true for
configurations with communication across servers ($w > 8, d < 8$ for 8-GPU
servers, e.g. $d=4$) where inter-stage all-to-all communication is cross-node
and more expensive.

\textbf{Compute-Communication Ratio:} Increasing the pipeline depth decreases
the amount of computation in each pipeline stage while keeping the number of
bytes communicated between stages constant. This makes the pipeline more
communication-bound, decreasing throughput.

\textbf{Maximum Per-GPU Microbatch Size:} Increasing the pipeline depth
increases the maximum microbatch size that fits in GPU memory. This leads to
possibly higher arithmetic intensity and throughput. In
Figure~\ref{fig:throughput_vs_batch_size}, we show throughput for two
microbatch sizes for the $d=8$ configuration; the larger microbatch size
($b=32$) has higher throughput. Smaller pipeline depths cannot fit large microbatch sizes.

\textbf{Maximum Model Size:} Deeper pipelines support the training of larger
models.  We show the empirically measured maximum model size that can be trained with \tbw using different values of $d$ in
Figure~\ref{fig:memory_size_vs_configuration}.

These observations illustrate the complexity in picking a configuration. For
example, increasing pipeline depth leads to two effects (decreased
compute-communication ratio within the pipeline and increased arithmetic
intensity) that have opposing effects on throughput. \system{}'s planner
automates this process for each combination of model, batch size, and number of GPUs.

\subsection{Maximum Model Size Supported}

\begin{figure}[t!]
    \centering
        \includegraphics[keepaspectratio=1.0,width=0.78\columnwidth]{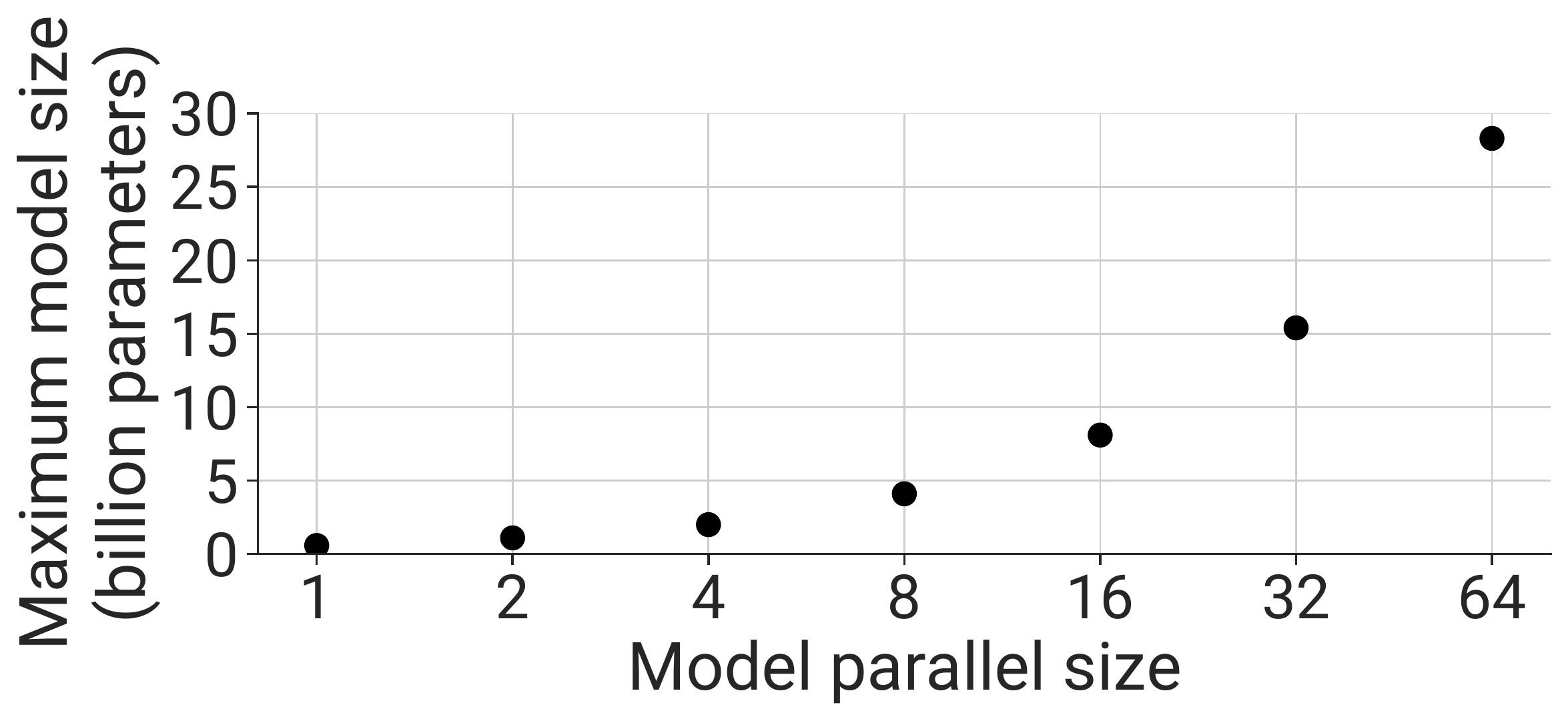}
    \caption{
        Maximum model size supported by various pipeline-parallel depths with
        64 16-GB V100 GPUs.
    }
    \label{fig:memory_size_vs_configuration}
    \vspace{-0.15in}
\end{figure}

Figure~\ref{fig:memory_size_vs_configuration} shows the empirically measured
maximum model size supported by various pipeline depths while using \tbw. As
can be seen in the figure, deeper configurations provide additional memory
capacity. \system{} is able to train models of up to almost 30 billion parameters
using 64 16-GB GPUs. As a point of comparison, Megatron-LM~\cite{shoeybi2019megatron}
was able to train a model with 8.3 billion parameters with 8 32-GB GPUs ($2\times$
more memory).

\section{Related Work and Discussion}
\label{sec:related_work}

In this section, we expand on work related to \system, and place \system's
speedups in context.

\textbf{Model Parallelism in Real Deployments.} NVIDIA used a custom
intra-layer model parallelism scheme in its Megatron
system~\cite{shoeybi2019megatron} to train a GPT-2 model with 8.3 billion
parameters on 64 32-GB V100 servers by parallelizing matrix multiplications
across multiple workers.
This approach can be combined with data parallelism. All-reductions are needed
to coalesce partial results produced on different GPUs, thus making training
communication-bound at high numbers of model partitions. In comparison, \system{} trades off additional memory footprint
(an extra weight version) for lower communication overhead
(20$\times$ faster training when using multiple multi-GPU servers on Amazon AWS with limited inter-node bandwidth).

\textbf{Pipeline Parallelism.} We discussed the existing approaches to pipeline
parallelism in \S\ref{sec:background}, and showed quantitative comparisons in
\S\ref{sec:evaluation_throughput_comparison}. \system{} trains large models up
to 3.2$\times$ faster than GPipe at low batch sizes, due to a lack of periodic
pipeline flushes, and lower memory footprint that allows more input
microbatches to be pushed into the pipeline. PipeDream cannot train these large
models. \system{}'s lower memory footprint does come with tradeoffs, however --
\system{} accumulates weight gradients over multiple microbatches,
increasing the minimum batch size that \system{} supports. Thus, for models
that only support very small batch sizes, \pdtbw{}, \pdflush, and GPipe,
which perform gradient \aggregation within the pipeline, may not be viable.

PipeMare~\cite{yang2019pipemare} uses asynchronous pipeline parallelism to
provide high throughput (no pipeline flushes) with asynchronous weight update
semantics.  PipeMare offers two theoretically-motivated techniques to ensure
good statistical efficiency. In contrast, \system{} and all the baselines we
compare against in the paper (traditional data parallel training, PipeDream,
GPipe), use synchronous execution where the weights used for computation during
forward propagation are the same as those used during backward propagation.
\system{}’s double buffered weight updates use a 1-stale gradient update that
does not require any hyperparameter tuning to generate comparable results. PipeMare also does not describe how computation
should be partitioned among the available workers.

\textbf{Memory-Saving Optimizations.} A rich line of work attempts to decrease
the memory footprint of DNN training.  Gist~\cite{jain2018gist} employs
lossless and lossy layer-specific encoding schemes to compress stashed
activations.
Systems such as Checkmate~\cite{mlsys2020_196} systematically determine when activation recomputation~\cite{chen2016training,
griewank2000algorithm}
should be performed.  DeepSpeed~\cite{rajbhandari2019zero} partitions optimizer
state over data-parallel replicas instead of replicating it, using a technique
called ZeRO.  Such orthogonal optimizations can be combined and incorporated in
\system{}.

\textbf{Planning Algorithms.} PipeDream, DAPPLE~\cite{fan2021dapple}, and
FlexFlow~\cite{flexflow} use planning algorithms to partition operator graphs
over multiple accelerators to maximize throughput. Unfortunately, these
planners do not exploit the repetitive nature of modern transformer-based
models. For example, PipeDream's planner explores $O(n^3m^2)$ configurations
(assuming $n$ layers in the model and $m$ workers). Furthermore, these
planners do not consider the effect of memory-saving optimizations, which
are critical for training large models efficiently (e.g., always applying
activation recomputation can make the system $1.33\times$ slower). \system's
planner, on the other hand, performs an exhaustive search of a \emph{much
reduced} search space since it only considers \emph{parallel pipelines} (all
possible $(w, d)$ pairs with $m$ workers is $O(m^2)$). Given this small
number of explored configurations, Bagpipe’s planner takes a fraction of
a second with a closed-form cost model; PipeDream's partitioning
algorithm with the same cost model takes about 30 minutes for large models.

\section{Conclusion}

In this work, we proposed and implemented \system{}, a system for
memory-efficient pipeline-parallel training that achieves high throughput, low
memory footprint, and data parallelism-like semantics through a novel weight
update double buffering strategy called \tbw. \system{} also uses a planner to
determine how to partition a model's operator graph over training resources in
a memory-aware way. \system{} accelerates the training of models with billions
of trainable parameters by up to 20$\times$ compared to model-parallel baselines, and by up to 3.2$\times$
compared to GPipe, on commodity hardware.

\section*{Acknowledgements}

We thank the anonymous reviewers, Aditya Grover, Paroma Varma, members of FutureData, and our colleagues at MSR for their feedback that improved this work.
We thank MSR for their generous support of Deepak's internship, and for resources to develop and evaluate \pdtbw.
This research was also supported in part
by affiliate members and other supporters of the Stanford DAWN project--Ant Financial, Facebook, Google, Infosys, NEC, and VMware--as well
as Northrop Grumman, Amazon Web Services,
Cisco, NSF Graduate Research Fellowship grant DGE-1656518, and the NSF CAREER grant CNS-1651570. Any opinions,
findings, and conclusions or recommendations expressed in this material
are those of the authors alone.

\bibliography{paper}
\bibliographystyle{icml2021}

\clearpage
\appendix
\section{Planner, Additional Details}
\label{sec:planner_appendix}

For every possible configuration of width and depth, \system{}'s planner
explores the benefit of pipelining and each space-saving optimization.  For
example, with activation recomputation as a target memory-savings optimization,
\system{} considers three possible executions:
\begin{itemize}[itemsep=1pt,topsep=0pt,leftmargin=*]
    \item Model and data parallelism without pipelining (with the largest per-GPU
          microbatch size that fits in memory).
    \item Hybrid parallelism with pipelining and without activation recomputation (all
    required weight versions and activation stashes in memory for in-flight microbatches).
    \item Hybrid parallelism with pipelining and recomputation.
\end{itemize}

\system{}'s planner estimates the throughput and memory footprint of each of these possible executions using a cost model.
\system{}'s planner then tries to find the configuration with highest throughput that also fits in main device memory of the accelerators used (memory capacity provided as input).
In this section, we show one such cost model for throughput and
memory.

\subsection{Closed-Form Cost Functions}

In our experiments, we used profile-based cost functions that run configurations end-to-end for a couple of hundred iterations.
However, performance of different parallel configurations can also be estimated
using closed-form expressions that use more fine-grained profile information
(e.g., time and memory footprint of each transformer block). We present one such cost model here.

\subsubsection{\textsc{throughput(.)} Cost Function} \label{sec:performance}

The throughput of various hybrid-parallel setups with and without pipelining
can be modeled using the times of forward and backward passes obtained from a
simple profiling step.  Let $b$ be the largest per-GPU microbatch size without
additional weight and activation versions, and $b'$ be the largest per-GPU
microbatch size that can fit on the device when multiple versions are needed
($b' \leq b$).  As before, $w$ and $d$ are the pipeline width and depth.

Let $T^{\text{comp}}_i(b, w, d)$ represent the compute time of stage $i$ with a
per-GPU microbatch size $b$, $T^{\text{comm}}_{i\rightarrow j}(b, w, d)$
represent the communication time of activations and gradients between stages
$i$ and $j$ with microbatch size $b$, and $T^{\text{comm}}_i(b, w, d)$
represent the communication time of exchanging gradients between $w$ replicas
of stage $i$ with microbatch size $b$. We assume that the global batch size used is $B$. With pipeline width $w$ and microbatch size $b$, data-parallel communication is required every $m(b) = B / (w \cdot b)$ microbatches.

Then, without pipelining, each microbatch of size $b$ takes the following
computation time, $t$:
\begin{align*}
t = \sum_i \max (& T^{\text{comp}}_i(b, w, d) + \sum_j T^{\text{comm}}_{j\rightarrow i}(b, w, d), \\
                 & \dfrac{1}{m(b)} \cdot T^{\text{comm}}_i(b, w, d))
\end{align*}
With pipelining, computation of different stages can be overlapped. A microbatch of size $b'$ can then be processed every $t$ seconds,
where $t$ is given by the expression:
\begin{align*}
t = \max_i \max (& T^{\text{comp}}_i(b', w, d) + \\
                 & \sum_j T^{\text{comm}}_{j\rightarrow i}(b', w, d), \\
                 & \dfrac{1}{m(b')} \cdot T^{\text{comm}}_i(b', w, d))
\end{align*}
With activation recomputation, the number of floating point operations
increases, since forward passes need to be repeated to recompute the activation
stashes needed in the backward pass. We use a constant multiplier
$c^{\text{extra}}$ to represent this.  $c^{\text{extra}} = 4/3$ is a reasonable
value for this constant, since the backward pass typically takes twice as long
as the forward pass. $c^{\text{extra}}$ can also be measured empirically.
Arithmetic intensity might also increase, which is captured by
$T_i^{\text{comp}}(.)$ being a function of the microbatch size $b$.
Communication time remains unchanged from before.  Every $b$ inputs can now be
processed in time $t$, where $t$ is given by,
\begin{align*}
t = \max_i \max (& c^{\text{extra}} \cdot T^{\text{comp}}_i(b, w, d) + \\
                 & \sum_j T^{\text{comm}}_{j\rightarrow i}(b, w, d), \\
                 & \dfrac{1}{m(b)} \cdot T^{\text{comm}}_i(b, w, d) )
\end{align*}

The throughput in samples per second of each of these setups is then the
corresponding per-GPU microbatch size ($b$ or $b'$) divided by $t$.

\textbf{Estimating $T^{\text{comp}}(.)$.} $T^{\text{comp}}_i(b, w, d)$ is
the compute time of stage $i$ with per-GPU microbatch size $b$, and can be
computed by summing up the forward and backward pass times of all blocks within
the stage. If the depth of the pipeline is $d$ and the total number of blocks
in the model is $B$, then the total number of blocks in a given stage is $B/d$.
Forward and backward pass times for each stage can be estimated by profiling
100--200 iterations of training.

\textbf{Estimating $T^{\text{comm}}(.)$.} Communication times can be
similarly modeled. Let the size of the associated parameter with $B$ total
blocks be $|W|$, and the size of the block's input and output activations be
$|A^{\text{inp.+out.}}(b)|$. With a pipeline of depth $d$, each pipeline stage
has $1/d$ of the total model parameters.

The time to communicate activations across stages can be computed as (factor of
2 for gradients in the backward pass), $$T^{\text{comm}}_{i\rightarrow j}(b, w,
d) = \frac{2|A^{\text{inp.+out.}}(b)|\cdot \mathbb{I}(d >
1)}{\text{bwdth}_{\text{depth}}(d)}$$

The time to communicate weight gradients across stage replicas can be computed
similarly given a bandwidth function $\text{bwdth}_{\text{width}}(w)$, and the
number of bytes communicated during all-reduce. The number of byes communicated
in an all-reduction can either be explicitly measured, or estimated using a
closed-form expression~\cite{narayanan2019pipedream}.

$\text{bwdth}_{\text{depth}}(d)$ and $\text{bwdth}_{\text{width}}(w)$ represent
the bandwidths for inter-stage and intra-stage communication. These bandwidth
functions can respect hierarchical network topologies. For example, if $w$ is
less than the number of workers in a single server, communication can be
performed entirely within a server, using the higher intra-server bandwidth.
$$
\text{bwdth}_{\text{width}}(w) =
    \begin{cases}
    B_\text{high} \text{ if } w < \text{number of GPUs in server} \\
    B_\text{low} \text{ otherwise}
    \end{cases}
$$

\begin{algorithm}[tb]
   \caption{Partitioning Algorithm}
   \label{alg:partitioning_algorithm}
\begin{algorithmic}
   \STATE {\bfseries Input:} Model $m$, memory capacity $M$, $m$'s associated search function $\textsc{search}(.)$, $m$'s associated throughput cost function $\textsc{throughput}(.)$, $m$'s memory footprint cost function $\textsc{memory}(.)$, maximum safe batch size $B$.
   \STATE {\bfseries Return:} Optimal width and depth $w^{\text{opt}}$ and $d^{\text{opt}}$, optimal per-GPU microbatch size $b^{\text{opt}}$, boolean whether activations should be recomputed $r^{\text{opt}}$, optimal degree of gradient \aggregation $g^{\text{opt}}$.
   \STATE
   \STATE Initialize $t^{\text{max}} = 0, w^{\text{opt}} = \text{NULL}, d^{\text{opt}} = \text{NULL}$
   \FOR{$w=1$ {\bfseries to} $N$}
   \FOR{$d=1$ {\bfseries to} $N/w$}
   \STATE\textit{// For given width $w$, depth $d$, and batch size $B$, find optimal microbatch size and whether activation recomputation should be performed.}
   \STATE $b, r = m.\textsc{search}(w, d, B)$
   \STATE
   \STATE $t = m.\textsc{throughput}(w, d, b, r)$
   \IF{$m.\textsc{memory}(w, d, b, r) > M$}
   \STATE \textbf{continue}
   \ENDIF
   \IF{$t > t^{\text{max}}$}
   \STATE $t^{\text{max}} = t, w^{\text{opt}} = w, d^{\text{opt}} = d, b^{\text{opt}} = b, r^{\text{opt}} = r$
   \ENDIF
   \ENDFOR
   \ENDFOR
   \STATE $g^{\text{opt}} = B / (N \cdot b^{\text{opt}})$ \textit{   // To reach batch size $B$.}
\end{algorithmic}
\end{algorithm}

\subsubsection{\textsc{memory(.)} Cost Function}
The memory footprint can similarly be modeled using the sizes of activations
and weights obtained from a profiling step.  Let the total size of the weight
parameters for the entire model be $|W|$, let the total size of the
activations given a microbatch size $b$ for the entire model be $|A^\text{total}(b)|$, and let
the size of the input activations for a single stage be $|A^\text{input}(b)|$. With a
pipeline of $d$ stages, each pipeline stage has weight parameters of size
$|W|/d$, and activations of size $|A^\text{total}(b)|/d$.

\textbf{Without Activation Recomputation.} As discussed in \S 3.1,
\tbw maintains 2 different versions of the weight parameters. \system{} also
maintains $d$ versions of activations (the total number of in-flight
activations). This means the total \system{} memory footprint is:
$$\frac{2|W|}{d} + \frac{d|A^\text{total}(b)|}{d} + d|A^\text{input}(b)|.$$

\textbf{With Activation Recomputation.} With activation recomputation, the
total number of activation versions in GPU memory at any point in time is 1.
This means that the \system{} memory footprint with $d$ stages is:
$$\frac{2|W|}{d} + \frac{|A^\text{total}(b)|}{d} + d|A^\text{input}(b)|.$$

\subsection{Partitioning Algorithm}

We show pseudocode for the full partitioning algorithm in Algorithm~\ref{alg:partitioning_algorithm}.
\section{Evaluation, Additional Graphs}

In this section, we present additional results we could not fit in the main
paper due to space.

\subsection{Throughput and Memory Footprint with BERT Models}
\label{sec:evaluation_bert}

\begin{figure}[t!]
    \centering
        \begin{subfigure}[c]{\columnwidth}
            \centering
            \includegraphics[keepaspectratio=1.0,width=0.82\columnwidth]{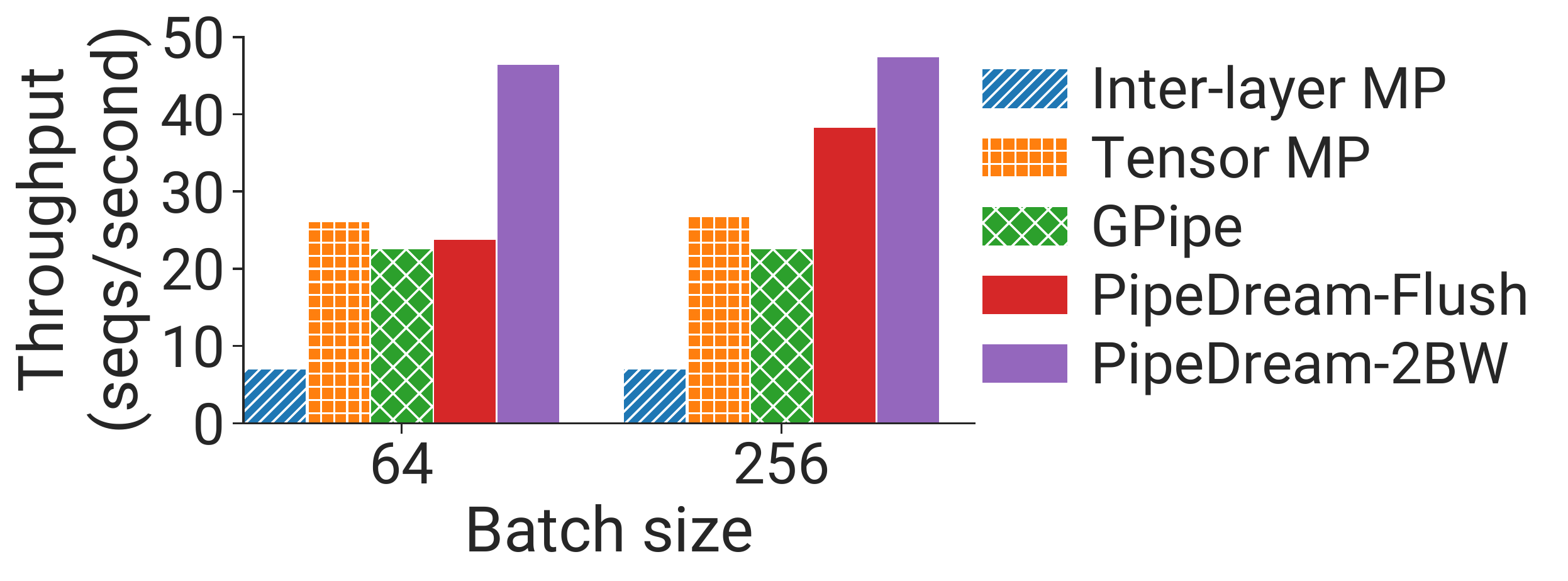}
            \vspace{-0.1in}
            \caption{BERT, 2.2B, 8-way model parallelism (8$\times$V100s).}
        \end{subfigure}
        \begin{subfigure}[c]{\columnwidth}
            \centering
            \includegraphics[keepaspectratio=1.0,width=0.82\columnwidth]{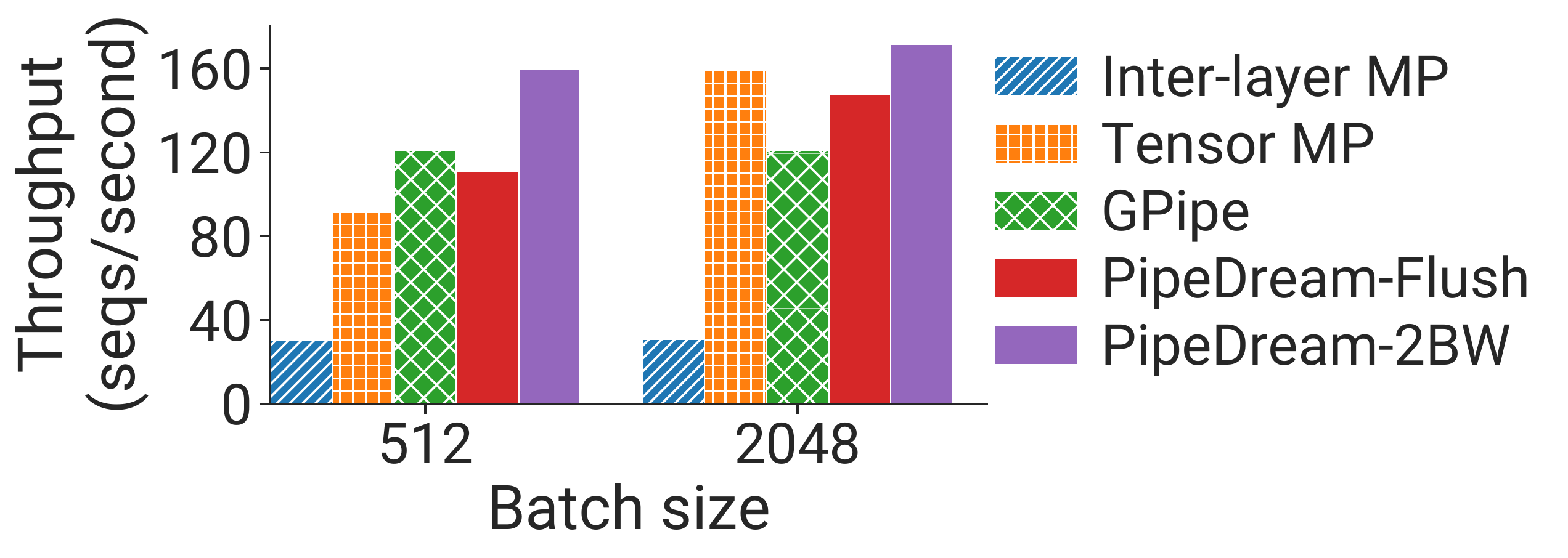}
            \vspace{-0.1in}
            \caption{BERT, 2.2B, 8-way model parallelism (64$\times$V100s).}
        \end{subfigure}
        \begin{subfigure}[c]{\columnwidth}
            \centering
            \includegraphics[keepaspectratio=1.0,width=0.82\columnwidth]{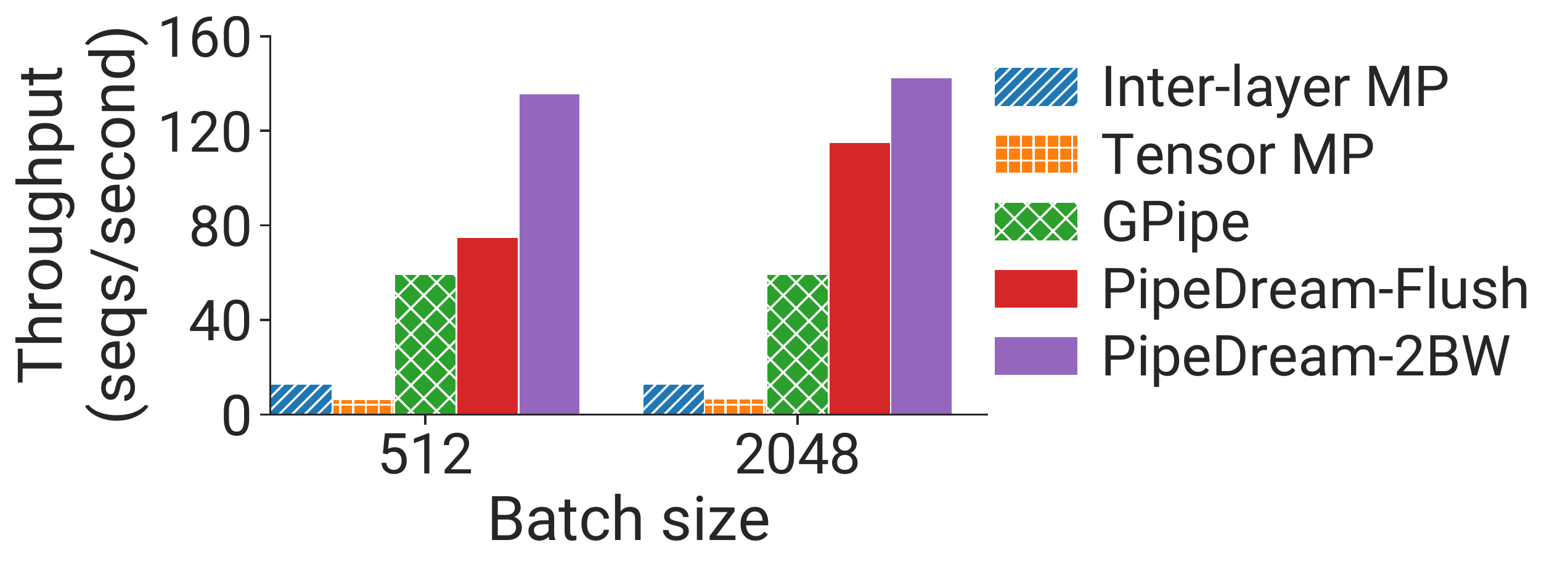}
            \vspace{-0.1in}
            \caption{BERT, 3.8B, 16-way model parallelism (64$\times$V100s).}
        \end{subfigure}
    \caption{
       Throughput of various systems for different batch sizes for
       BERT models. Results are shown with a single 8$\times$V100 server, and
       with eight 8$\times$V100 servers (with 16GB).
    }
    \label{fig:throughput_vs_other_pipelining_approaches_bert}
\end{figure}

\begin{figure}[t!]
    \centering
    \includegraphics[keepaspectratio=1.0,width=0.84\columnwidth]{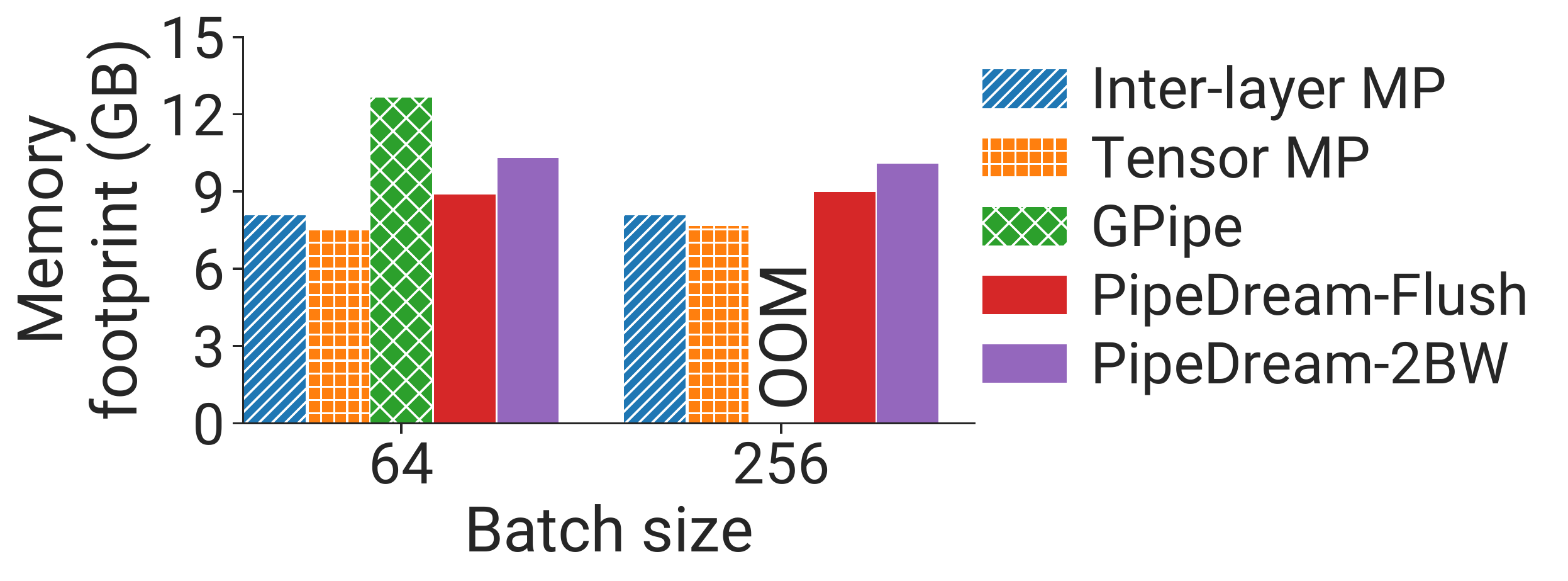}
    \vspace{-0.1in}
    \caption{
        Worst-case memory footprint (in GB) of various systems
        with 8 V100 GPUs for a BERT model with 2.2B parameters.
    }
    \label{fig:memory_footprint_bert}
\end{figure}

We also ran \system{} on two BERT models: one with 2.2 billion parameters, and
another with 3.8 billion parameters.
Figure~\ref{fig:throughput_vs_other_pipelining_approaches_bert} compares
\system{}'s throughput against the same baselines as before, and
Figure~\ref{fig:memory_footprint_bert} compares \system{}'s memory footprint
for these BERT models. We see that results are similar to GPT. One point of
difference is that GPipe does not run out of memory at the batch size
of 64 (for GPT, only a batch size of 32 fits in memory, leading to a larger pipeline bubble); however, GPipe still has higher memory footprint compared to
all other baselines.

\begin{figure}[t!]
    \centering
        \begin{subfigure}[c]{0.49\columnwidth}
            \includegraphics[keepaspectratio=1.0,width=\columnwidth]{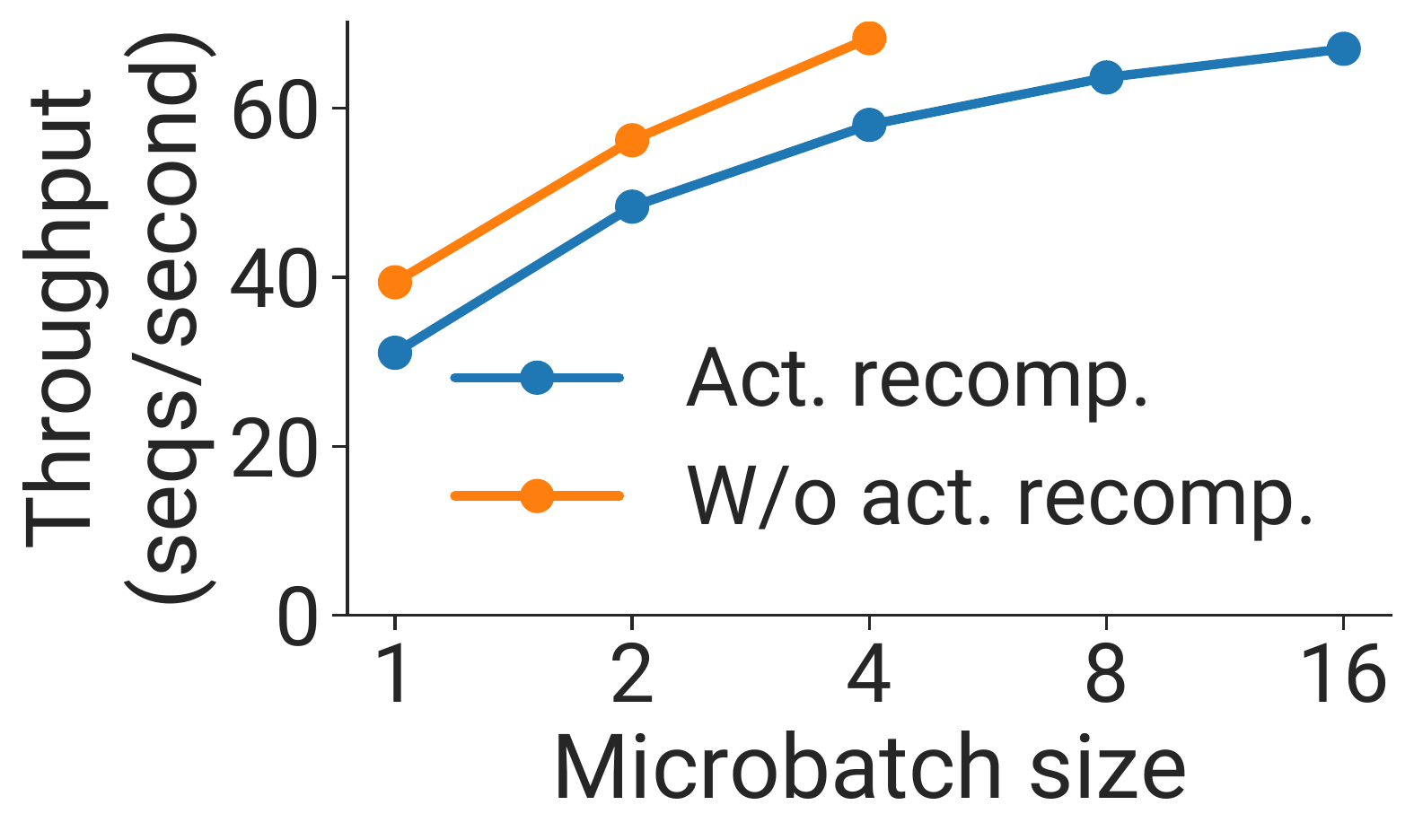}
            \caption{GPT, 1.3B.}
            \label{fig:throughput_vs_microbatch_size_1.3b}
        \end{subfigure}
        \begin{subfigure}[c]{0.49\columnwidth}
            \includegraphics[keepaspectratio=1.0,width=\columnwidth]{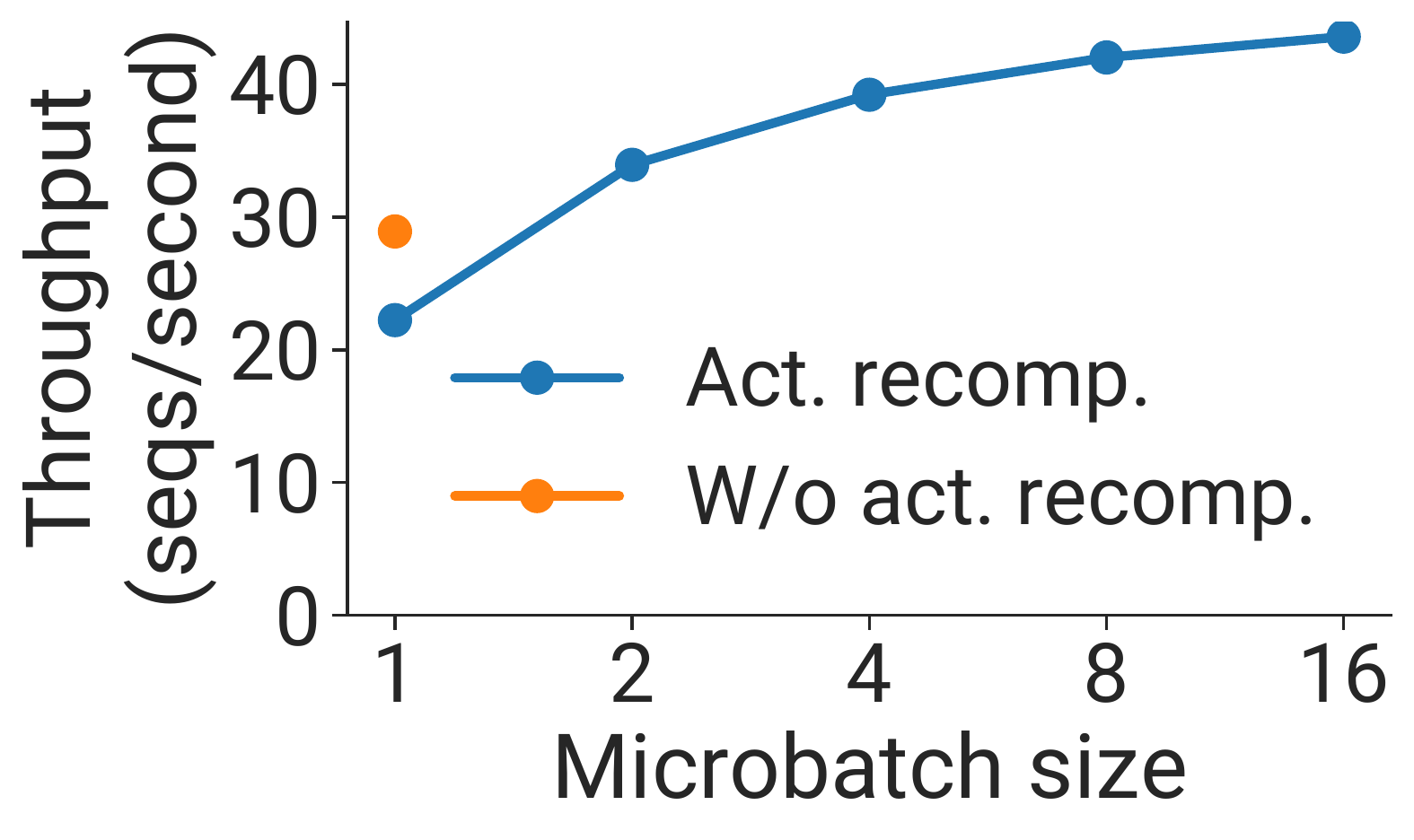}
            \caption{GPT, 2.2B.}
            \label{fig:throughput_vs_microbatch_size_2.2b}
        \end{subfigure}
    \caption{
       Throughput of $(1, 8)$ \system{} configurations vs. per-GPU microbatch size
       for GPT models using a maximum sequence length of 512 and 8 16-GB-V100 GPUs,
       with and without activation recomputation. Activation recomputation helps increase
       the maximum per-GPU microbatch size that fits, especially for larger models,
       leading to higher throughput in some cases.
    }
    \label{fig:throughput_vs_microbatch_size}
\end{figure}

\subsection{Impact of Activation Recomputation}
\label{sec:evaluation_bagpipe_configuration_comparison_activation_recomputation}

Figure~\ref{fig:throughput_vs_microbatch_size} shows the effect of activation
recomputation on throughput for various GPT models. For a given per-GPU
microbatch size, recomputation introduces overhead (capped at $33\%$ since the
backward pass takes twice as long as the forward pass for most operators).
However, recomputation allows for a larger per-GPU microbatch to fit on the
worker, sometimes leading to higher throughput than without activation
recomputation: activation recomputation leads to higher throughput in
Figure~\ref{fig:throughput_vs_microbatch_size_2.2b}, but not in
Figure~\ref{fig:throughput_vs_microbatch_size_1.3b}.  In the extreme case (not
pictured), recomputation makes it possible to train large models by reducing
peak memory footprint of training, at the cost of extra compute operations due to an extra forward pass.

\end{document}